\documentclass[11pt]{article}

\usepackage[final]{acl}

\usepackage{times}
\usepackage{latexsym}
\usepackage{amsmath}
\usepackage{amssymb}
\usepackage{booktabs}
\usepackage{multirow}
\usepackage{makecell}
\usepackage[table]{xcolor}

\usepackage[T1]{fontenc}

\usepackage[utf8]{inputenc}

\usepackage{microtype}

\usepackage{inconsolata}

\usepackage{graphicx}

\title{PN-Mix: Pair-Norm Guided Mixed-Precision Key Cache Quantization for Efficient Long-Output LLM Inference}

\author{
 \textbf{Gil Jae Song} \quad \textbf{Sungjoo Yoo}\thanks{Indicates corresponding author}
\\
Seoul National University
\\
\texttt{jkoi632@snu.ac.kr, sungjoo.yoo@gmail.com}
}

\begin{document}
\maketitle
\begin{abstract}
Long-output inference in large language models suffers from severe KV cache memory and bandwidth bottlenecks. 
While KV cache quantization alleviates this issue, low-bit quantization (e.g., 2-bit) severely degrades model accuracy. Recent mixed-precision key-cache quantization methods mitigate this degradation by allocating higher precision to important key channels, but estimate channel saliency using only key statistics.
Through attention logit perturbation analysis at the RoPE channel-pair level, we show that quantization errors in channel pairs with large query--key pair-norm products contribute significantly to logit variation. 
Based on this insight, we propose PN-Mix, a mixed-precision key-cache quantization method for RoPE-based decoder-only LLMs, which allocates higher precision to salient channel pairs identified once during the prefill stage and reused throughout decoding.
Across standard reasoning, code-generation, and AIME benchmarks, PN-Mix with 2.26-bit average precision achieves comparable or better average accuracy than prior mixed-precision methods using 2.32--2.39 bits. 
Furthermore, in our Pareto sweep, PN-Mix degrades gracefully down to 1.5 bits, whereas the compared baselines exhibit sharper performance drops below 2.25 bits.
\end{abstract}

\begin{figure}[t]
\centering
\includegraphics[width=\linewidth]{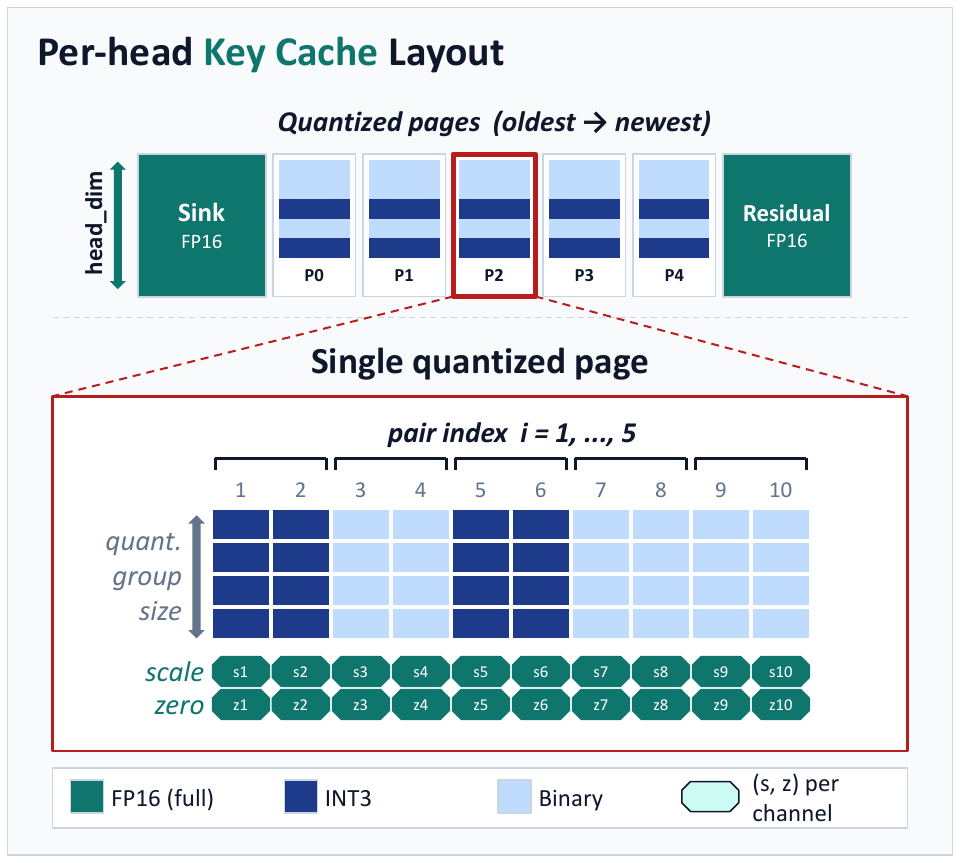}

\caption{
\textbf{Post-RoPE key-cache layout of PN-Mix.}
Salient RoPE channel pairs selected during prefill are stored in higher precision across quantized pages, while the remaining pairs use lower precision.}
\label{fig:method_overview}
\end{figure}

\section{Introduction}
Long-output inference has become a central capability of large language models (LLMs), enabling applications that require extended generation such as code generation and complex problem solving.
However, long-output decoding exposes a severe systems bottleneck in the key-value (KV) cache, which stores previous keys and values but grows linearly with sequence length and batch size.
This growth increases memory pressure, reducing the feasible batch size for serving and limiting the maximum generation length.
For example, Qwen3-8B in FP16 runs out of memory on a single A100-40GB GPU when the generation length exceeds 4K at batch size 32, or when the batch size exceeds 20 at an 8K generation length (Figure~\ref{appendix:peak-mem-fig}).

KV cache quantization alleviates this bottleneck by storing cached keys and values at lower precision while retaining all token positions and the original cache structure.
Yet this compression is inherently lossy: aggressive low-bit quantization (e.g., 2-bit) can introduce errors that substantially degrade reasoning accuracy over long chains of thought~\citep{liu2025quantization}.
Thus, effective KV cache quantization requires not only lowering precision, but also allocating the limited bit budget to the most critical components.

Recent mixed-precision methods allocate higher precision to salient key channels, motivated by the greater quantization sensitivity of keys than values~\citep{li2025kvtuner, xia2025kittyaccurateefficient2bit, hariri-etal-2026-quantize}. 
However, their saliency estimates rely on key statistics alone and operate at the individual-channel level, overlooking query--key interactions and the channel-pair structure induced by RoPE.

In this work, we propose PN-Mix, a pair-norm guided mixed-precision key-cache quantization method designed for RoPE-based decoder-only LLMs.
Our key observation is that key-cache quantization should be guided by its effect on attention logits, rather than by key statistics alone.
Through attention-logit perturbation analysis at the RoPE channel-pair level, we show that quantization errors in channel pairs with large query--key pair-norm products can contribute substantially to logit variation.
PN-Mix therefore assigns higher precision to such salient RoPE channel pairs.
We further find that salient channel pairs estimated from prefill tokens are predictive of full-sequence saliency, enabling PN-Mix to identify them once during prefill and reuse them throughout decoding.
As a result, PN-Mix remains input-adaptive without offline calibration and avoids repeated saliency recomputation during decoding.

Overall, this work makes three main contributions:
\begin{itemize}
    \item We derive a query-aware RoPE channel-pair saliency criterion based on query--key pair-norm products from attention-logit perturbation analysis.
    \item We propose PN-Mix, a mixed-precision key-cache quantization method that input-adaptively selects salient RoPE channel pairs once during prefill and reuses the resulting selection throughout decoding.
    \item We show that PN-Mix improves long-output inference by reducing average key-cache precision while preserving accuracy and avoiding decoding-time saliency recomputation.
\end{itemize}


\section{Background}
\label{sec:backgrounds}

We provide a brief overview of the attention and online KV-cache quantization setup used in this work.
For a decoder-layer hidden state $\mathbf{X} \in \mathbb{R}^{T \times D}$ with sequence length $T$ and hidden dimension $D$, Grouped Query Attention (GQA)~\citep{ainslie2023gqa} projects $\mathbf{X}$ into $Q \in \mathbb{R}^{H \times T \times d}$ and $K,V \in \mathbb{R}^{H^{kv} \times T \times d}$, where $H$, $H^{kv}$, and $d$ denote the number of query heads, KV heads, and the head dimension.
RoPE~\citep{su2024roformer} gives $\tilde{Q}_{h',t}=Q_{h',t}R_t$ and $\tilde{K}_{h,t}=K_{h,t}R_t$, where $R_t$ is a position-dependent block-diagonal rotation over adjacent channel pairs.
With causal masking, the attention distribution $A_{h,t}$ and output $O_{h,t}$ at query position $t$ are computed over positions $1,\ldots,t$ as
\begin{equation}
\begin{aligned}
A_{h,t} = \mathrm{softmax}\!\left(\tfrac{\tilde{Q}_{h',t}\tilde{K}_{g(h'),1:t}^{\top}}{\sqrt{d}}\right), \\ 
O_{h,t}=A_{h,t}V_{g(h'),1:t}.
\end{aligned}
\label{eq:attention}
\end{equation}
where $g(h')$ maps query head $h'$ to its shared KV head $h$.
During autoregressive decoding, previously computed $\tilde{K}$ and $V$ are stored as the KV cache and reused at later steps.
For batch size $B$ and number of layers $L$, the KV cache memory scales as $\mathcal{O}(B L H^{kv} T d)$, motivating KV-cache quantization for long-output inference.

We adopt asymmetric group quantization as the base quantization scheme.
For a group $X \in \mathbb{R}^{G}$, $b$-bit asymmetric quantization stores $X_q \in \{0,\ldots,2^b-1\}^{G}$ with scale $s_X$ and zero-point $z_X$, and reconstructs $\hat{X}=s_X X_q+z_X$:
\begin{equation}
\begin{aligned}
s_X &= \tfrac{\max X - \min X}{2^b - 1}, \quad z_X = \min X, \\
X_q &= \mathrm{clamp}\!\left(\left\lfloor \tfrac{X - z_X}{s_X} \right\rceil, 0, 2^b-1\right).
\end{aligned}
\label{eq:quantize}
\end{equation}
Following common KV-cache quantization practice~\citep{kivi}, we apply different grouping strategies to keys and values: keys are grouped per channel to handle channel-wise outliers, whereas values are grouped per token to align with the attention weighted-sum operation in Eq.~\ref{eq:attention}.
In online decoding, newly generated $\tilde{K}_{h,t}$ and $V_{h,t}$ are kept in an FP16 residual buffer of size $N$.
Once the buffer is full, it is quantized and appended to the quantized cache; during attention, the dequantized cache is concatenated with the FP16 residual buffer.


\section{Related Work}

Prior work has explored several directions for reducing KV cache memory, including token eviction~\citep{zhang2023h2o, li2024snapkv, cai2026rkv}, low-rank approximation~\citep{chang2025palu, tale2025}, merging~\citep{pmlr-v235-zhang24n, wang2024model}, pruning~\citep{xu2025think, zhang-etal-2025-leank}, and quantization.
Among them, KV cache quantization reduces memory by lowering the representation precision of cached tensors while retaining the cache structure.
We focus on KV cache quantization, which can be broadly divided into uniform-precision and mixed-precision approaches.

\subsection{Uniform-Precision Quantization}

Uniform-precision KV cache quantization assigns the same bit-width to all quantized cache elements.
Prior methods reduce low-bit quantization error through key/value-specific grouping~\citep{kivi}, distribution-aware quantization~\citep{hooper2024kvquant}, residual correction~\citep{pmlr-v262-kang24a}, outlier-token preservation~\citep{su2025accurate}, Walsh--Hadamard channel rotations~\citep{10.24963/ijcai.2025/690}, projection-based 1-bit sign quantization~\citep{zandieh2024qjl}, and polar-coordinate quantization of RoPE channel pairs~\citep{wu2025polarquant}.
However, because they assign the same bit-width regardless of cache-element sensitivity, uniform-precision methods may insufficiently control errors in important components, which can accumulate over long chains of thought in reasoning models~\citep{liu2025quantization}.

\subsection{Mixed-Precision Quantization}

Mixed-precision quantization instead allocates a limited bit budget according to estimated importance.
Existing methods differ in allocation granularity, including token-level~\citep{yang2024tokenleftbehindreliable}, layer-level~\citep{li2025kvtuner, liu2026pmkvq}, and channel-level allocation.
Since key-cache quantization importance is highly non-uniform across channels due to channel-wise outliers~\citep{kivi, hooper2024kvquant}, recent channel-level methods focus on assigning higher precision to salient key channels.
ChanMix~\citep{liao2026channelaware} uses an offline calibration set to identify retrieval-head channels, assigns them 4-bit precision, and allocates 3/2/1-bit precision to the remaining channels via min--max range-based clustering.
Kitty~\citep{xia2025kittyaccurateefficient2bit} dynamically re-estimates channel saliency for each quantization group using average key magnitude, assigning 4-bit precision to salient channels and 2-bit precision to the rest.

Although these methods improve over uniform precision, they estimate saliency from $\tilde{K}$ alone and assign precision at the individual-channel level.
This misses two structures that directly affect attention: attention logits are determined by the interaction between $\tilde{Q}$ and $\tilde{K}$, and RoPE couples channels into pairs.
In contrast, PN-Mix estimates saliency from query--key interactions at the RoPE channel-pair level and allocates mixed precision to the key cache accordingly.

\section{Preliminary Analysis}
\label{sec:preliminary}

To achieve accuracy-preserving $\sim$2.25-bit KV-cache compression for long-output inference, we investigate three design questions:
(1) which cache (key or value) to refine;
(2) how to define an importance score for assigning higher precision within the key cache; and
(3) how to exploit this importance score efficiently during long-output inference.
We examine these questions sequentially and use the resulting observations to guide our mixed-precision key-cache quantization design.

\subsection{Sensitivity of Key and Value Caches to Quantization}
\label{subsec:kv-sensitivity}

Under a 2.25-bit budget\footnote{Effective bit-width for 2-bit asymmetric quantization ($G=128$) with per-group FP16 scales and zero-points. Tokens in the FP16 residual buffer are excluded.}, prior work indicates the key cache is more sensitive to quantization than the value cache~\citep{li2025kvtuner, xia2025kittyaccurateefficient2bit}. 
We verify this by quantizing either $\tilde{K}$ or $V$ to 2.25 bits and comparing against an FP16 baseline on AIME datasets. 
We evaluate DeepSeek-R1-Distill-Llama-8B (R1-Llama-8B) and Qwen3-8B, generating up to 32K tokens per response. 
Accounting for attention sinks~\citep{xiao2024efficient}, we also evaluate variants preserving the initial 32 tokens in FP16, following prior work~\citep{xia2025kittyaccurateefficient2bit}. 

Table~\ref{appendix:kv-sink-compare} shows that $V$-only quantization causes less degradation than $\tilde{K}$-only quantization, even without sink-token protection. 
While preserving the first 32 tokens in FP16 improves both, $V$-only quantization approaches the FP16 baseline whereas $\tilde{K}$-only quantization still exhibits non-negligible degradation.

These results lead to two design choices.
First, sink tokens should be preserved in full precision for both caches.
Second, key-cache quantization should be refined more carefully than value-cache quantization under low-bit compression.

\begin{figure*}[t]
\centering
\includegraphics[width=\linewidth]{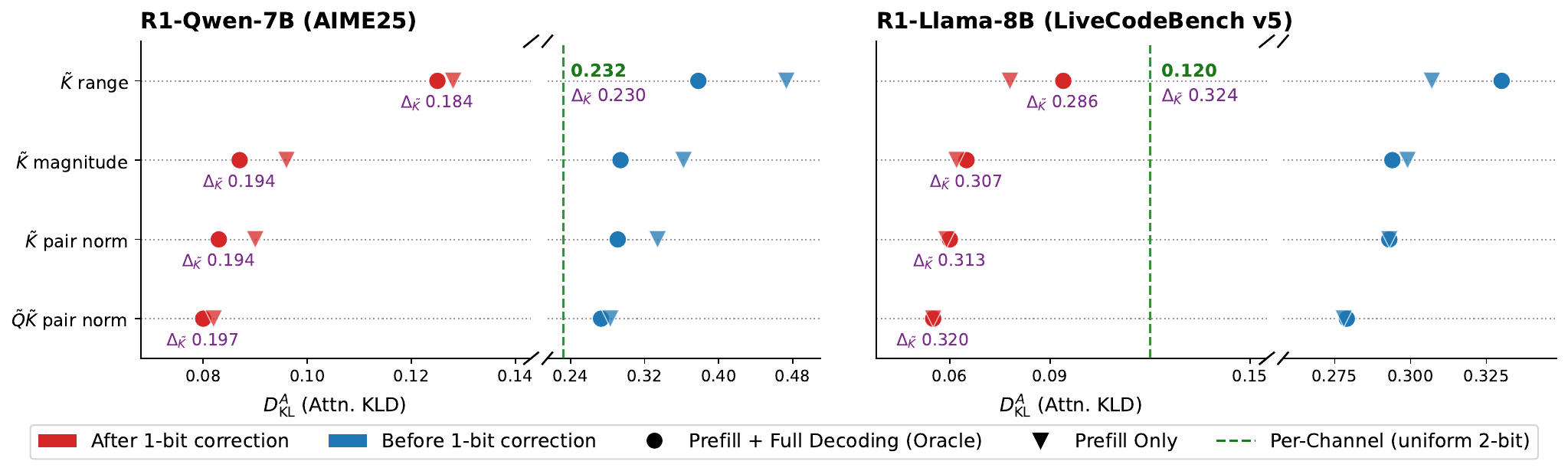}
\caption{\textbf{Simulation results under H3(50\%)L1 key-cache bit allocation.}
Each marker shows mean $D_{\mathrm{KL}}^{A}$; adjacent purple values show mean $\Delta_{\tilde{K}}$.
For each criterion, we pseudo-quantize $\tilde{K}$ with midpoint correction for 1-bit groups (Eq.~\ref{eq:1bit_correction}); $\Delta_{\tilde{K}}$ is measured on the reconstructed $\tilde{K}'$, while $D_{\mathrm{KL}}^{A}$ is computed from $A'_{h,t}$ recomputed using the corresponding $\tilde{Q}$.}
\label{fig:k_quant_simul}
\end{figure*}

\subsection{Designing an Attention-Aware Importance Score}
\label{subsec:channel-pair-importance}

Having identified the key cache as the more sensitive component under low-bit quantization, we next consider how to determine which parts of the key cache should receive higher precision. 
Existing mixed-precision key-cache quantization methods estimate channel importance using statistics derived only from the key cache. 
ChanMix~\citep{liao2026channelaware} uses the channel range, $\max_t \tilde{K}_{h,t,c}-\min_t \tilde{K}_{h,t,c}$, 
whereas Kitty~\citep{xia2025kittyaccurateefficient2bit} uses the average channel magnitude, $\mathbb{E}_t[|\tilde{K}_{h,t,c}|]$.

However, these key-only, channel-wise criteria do not explicitly account for how the key cache is used in attention.
After RoPE, adjacent key channels are coupled into channel pairs, and each key contributes to the attention logits through its interaction with the query.
Motivated by these observations, we decompose the attention-logit perturbation caused by post-RoPE key quantization at the RoPE channel-pair level.

For a fixed head $h$, we define the $i$-th query and key channel-pair vectors at query position $t$ and key position $s$ as
$\mathbf{q}^{t}_{i}:=(\tilde{Q}_{h,t,2i-1},\tilde{Q}_{h,t,2i})$ and
$\mathbf{k}^{s}_{i}:=(\tilde{K}_{h,s,2i-1},\tilde{K}_{h,s,2i})$.
For $s\leq t$, dropping the common scaling factor $1/\sqrt{d}$, the corresponding attention logit can be decomposed as
\begin{equation}
\tilde{Q}_{h,t}\tilde{K}_{h,s}^{\top}
=
\sum_{i=1}^{d/2}
\|\mathbf{q}^{t}_{i}\|_2
\|\mathbf{k}^{s}_{i}\|_2
\cos(\phi^{t,s}_{i})
\label{eq:logit_decomp}
\end{equation}
where $\phi^{t,s}_{i}$ denotes the angle between $\mathbf{q}^{t}_{i}$ and $\mathbf{k}^{s}_{i}$.
Let $\mathbf{k}'^{s}_{i}$ denote the reconstructed key pair after quantization.
We express the resulting changes in the key-pair-norm and angular term as
$\|\mathbf{k}'^{s}_{i}\|_2=\|\mathbf{k}^{s}_{i}\|_2+\epsilon^{s}_{i}$ and
$\cos(\phi'^{t,s}_{i})=\cos(\phi^{t,s}_{i})+\delta^{t,s}_{i}$,
respectively, and define the relative perturbation of the key-pair-norm as
$\rho_i^s=\epsilon_i^s/\|\mathbf{k}_i^s\|_2$.
The resulting difference between the original and quantized attention logits is
\begin{equation}
\begin{aligned}
&\tilde{Q}_{h,t}\tilde{K}_{h,s}^{\top}
-\tilde{Q}_{h,t}\tilde{K}'_{h,s}{}^{\top} \\
&= -\sum_{i=1}^{d/2}
\|\mathbf{q}_i^t\|_2\|\mathbf{k}_i^s\|_2
\left(
\delta_i^{t,s}
+\cos\phi_i^{t,s}\rho_i^s
+\rho_i^s\delta_i^{t,s}
\right)
\end{aligned}
\label{eq:logit_diff}
\end{equation}

In Eq.~\ref{eq:logit_diff}, the term in parentheses captures the perturbation induced by quantization, 
while the query--key pair-norm product scales its contribution to the attention-logit error. 
The exact perturbation terms, such as $\rho_i^s$ and $\delta_i^{t,s}$, depend on the reconstructed key and are therefore not available when the bit-width allocation is determined. 
We therefore use the query--key pair-norm product, which is directly available from the unquantized attention states, as an attention-aware proxy for the pre-quantization sensitivity of each RoPE channel pair.

Using this quantity directly as an importance score would require aggregating $\|\mathbf{q}^{t}_{i}\|_2\|\mathbf{k}^{s}_{i}\|_2$  over causal query--key position pairs $(t,s)$ with $s\leq t$.
Since our goal is to rank channel pairs for precision allocation rather than to reproduce the causal-joint score exactly, 
we use a separable approximation that summarizes the query and key pair-norms independently over the observed tokens. 
This avoids retaining the explicit causal pairing structure while preserving a compact score for channel-pair ranking.
Specifically, we define the importance score for channel pair $i$ in head $h$ as
\begin{equation}
\sigma_{h,i}
=
\mathbb{E}_{t}\!\left[\|\mathbf{q}^{t}_{i}\|_2\right]
\cdot
\mathbb{E}_{s}\!\left[\|\mathbf{k}^{s}_{i}\|_2\right]
\label{eq:importance_form}
\end{equation}

To assess the approximation, we compare it against a causal-joint score obtained by averaging $\|\mathbf{q}^{t}_{i}\|_2\|\mathbf{k}^{s}_{i}\|_2$ over all causal pairs $(t,s)$ with $s\leq t$.
As shown in Table~\ref{tab:causal_separable_sim}, Spearman rank correlation and top-$k\%$ selection overlap indicate that the separable formulation largely preserves the salient-pair ordering of the causal-joint score while avoiding explicit computation over all causal query--key pairs.

\begin{table}[t]
\centering
\small
\setlength{\tabcolsep}{3pt}
\begin{tabular}{l|ccc}
\toprule
\textbf{Model / Dataset}
& \makecell{\textbf{Spearman}\\$\boldsymbol{\rho}$}
& \makecell{\textbf{Top-10\%}\\\textbf{overlap}}
& \makecell{\textbf{Top-50\%}\\\textbf{overlap}} \\
\midrule
\makecell[l]{R1-Llama-8B /\\LiveCodeBench~v5}
& $0.996_{\pm 0.000}$
& $96.2_{\pm 0.3}$
& $97.9_{\pm 0.1}$ \\
\addlinespace
\makecell[l]{R1-Qwen-7B /\\AIME25}
& $0.990_{\pm 0.003}$
& $95.3_{\pm 0.6}$
& $96.2_{\pm 0.5}$ \\
\bottomrule
\end{tabular}
\caption{\textbf{Separable vs.\ causal-joint pair-norm score.}
The table reports the similarity in ranking and selection between $\sigma_{h,i}$ (Eq.~\ref{eq:importance_form}) and the causal-joint score computed over prefill states.
Metrics are averaged over layers and KV heads and reported as mean$_{\pm\mathrm{std}}$ across samples.}
\label{tab:causal_separable_sim}
\end{table}

Next, we evaluate whether the proposed importance score leads to more accuracy-preserving key-cache quantization than existing criteria.
The evaluation uses post-RoPE $\tilde{Q}$ and $\tilde{K}$ states collected from R1-Qwen-7B on AIME25 and R1-Llama-8B on LiveCodeBench~v5~\citep{jain2024livecodebench}.
Quantization quality is measured by two metrics:
the key-cache reconstruction error,
$\Delta_{\tilde{K}}:=\frac{1}{H^{kv}Td}\|\tilde{K}-\tilde{K}'\|_F^2$,
where $\tilde{K}'$ is the reconstructed key cache, and the attention-distribution KL divergence,
$D_{\mathrm{KL}}^{A}:=\mathbb{E}_{h,t}[D_{\mathrm{KL}}(A_{h,t}\,\|\,A'_{h,t})]$,
where $A'_{h,t}$ is recomputed using $\tilde{K}'$.

For comparison, we consider uniform 2-bit per-channel quantization and a mixed-precision setting, denoted H3(50\%)L1, that assigns 3 bits to the top 50\% of channels or channel pairs and 1 bit to the remainder. 
Bit allocation is determined by four importance criteria: the channel-wise $\tilde{K}$ range, the channel-wise $\tilde{K}$ magnitude, the $\tilde{K}$ pair norm $\mathbb{E}_{s}\!\left[\|\mathbf{k}^{s}_{i}\|_2\right]$, and the proposed $\tilde{Q}\tilde{K}$ pair norm $\sigma_{h,i}$. 
For the first two criteria, individual channels are ranked by importance, whereas the latter two rank channel pairs. 
In all cases, salient channels or channel pairs are selected using full-sequence oracle statistics. 
All methods use asymmetric per-channel quantization with group size $G=128$ (Eq.~\ref{eq:quantize}) under the same effective $\sim$2.25-bit budget. 
Figure~\ref{fig:k_quant_simul} summarizes the results, with full simulation details provided in Appendix~\ref{appendix:k-quant-simul}.

As shown in Figure~\ref{fig:k_quant_simul}, the proposed $\tilde{Q}\tilde{K}$ pair norm achieves the lowest $D_{\mathrm{KL}}^{A}$ among all compared mixed-precision criteria on both models, indicating better preservation of the attention distribution under the same effective bit budget.
Its improvement over the $\tilde{K}$ pair norm further indicates that incorporating query-side information provides additional benefit beyond accounting for the RoPE channel-pair structure alone.
Based on these results, we use $\sigma_{h,i}$ in Eq.~\ref{eq:importance_form} as the importance score for mixed-precision key-cache quantization.


\subsection{Efficient Importance Estimation for Long-Output Inference}
\label{subsec:eff-importance}

With $\sigma_{h,i}$ established as the importance score, the remaining challenge is to estimate it efficiently during long-output inference. 
ChanMix relies on offline profiling, eliminating online selection overhead but sacrificing input-adaptive channel allocation, 
whereas Kitty performs input-adaptive selection for each quantization group. 
However, this requires repeatedly identifying salient channels and storing a separate set of channel indices for each group. 
As decoding proceeds and more groups accumulate, the resulting index storage and loading overhead can grow substantially.

To retain input-adaptive importance while avoiding repeated salient-pair selection and index management during long decoding, 
we choose to compute $\sigma_{h,i}$ once from the prefill tokens and reuse the selected channel pairs throughout decoding.

For this prefill-only strategy to be effective, the prefill-based importance score should satisfy two properties.
First, it should be predictive: channel pairs identified as important during prefill should remain salient throughout subsequent decoding.
Second, it should be selective: the importance should be concentrated on a subset of channel pairs so that assigning higher precision to the selected pairs remains effective under a limited bit budget.

We first evaluate the predictiveness of prefill-based selection by comparing its quantization quality against a full-sequence oracle that computes importance scores using the complete sequence.
Using the same H3(50\%)L1 simulation setup as in \S\ref{subsec:channel-pair-importance}, Figure~\ref{fig:k_quant_simul} shows that the prefill-based $\tilde{Q}\tilde{K}$ pair norm achieves $D_{\mathrm{KL}}^{A}$ close to its full-sequence counterpart on both evaluated models.

To further understand why prefill-based selection remains effective, Appendix~\ref{appendix:ranking} compares the channel pairs identified from prefill with those that are salient during decoding. 
Among the evaluated criteria, the prefill-based $\tilde{Q}\tilde{K}$ pair norm yields the highest average Jaccard overlap with decoding-time salient pairs.
Moreover, Appendix~\ref{appendix:prefill_drift_decode} shows that this overlap does not systematically decrease as decoding progresses, including in long-generation and multi-topic settings. 

We next examine selectivity.
Approximately 97--99\% of the total importance is concentrated in the top 50\% of channel pairs (Figure~\ref{appendix:sigma_fig}); these values denote the median across layers, KV heads, and prompts.

Together, these results indicate that $\sigma_{h,i}$ can be computed once during the input prefill stage and that the selected channel pairs can be reused throughout long-output decoding.

\section{Methodology}
\label{sec:method}

This section presents PN-Mix, which integrates the channel-pair importance score $\sigma_{h,i}$ from \S\ref{subsec:channel-pair-importance} into the online KV-cache quantization setup in \S\ref{sec:backgrounds}.
As shown in Figure~\ref{fig:method_overview}, PN-Mix assigns mixed precision to key-cache channel pairs according to $\sigma_{h,i}$ while keeping sink tokens and the residual buffer in FP16.

\subsection{Prefill-Stage Salient Channel-Pair Prediction}
\label{subsec:prefill-selection}

During prefill, PN-Mix computes $\sigma_{h,i}$ from the prefill $\tilde{Q}$ and $\tilde{K}$ states for each layer, KV head $h$, and channel pair $i$.
Under GQA, each KV head is shared by multiple query heads, so we average the query-side scores over the query heads sharing $h$:
\begin{equation}
\begin{aligned}
\bar{q}_{h', i}
&= \mathbb{E}_{t}\!\left[
\left\| (\tilde{Q}_{h', t, 2i-1},\, \tilde{Q}_{h', t, 2i}) \right\|_2
\right], \\
\bar{k}_{h, i}
&= \mathbb{E}_{t}\!\left[
\left\| (\tilde{K}_{h, t, 2i-1},\, \tilde{K}_{h, t, 2i}) \right\|_2
\right], \\
\sigma_{h,i}
&= \tfrac{1}{|g^{-1}(h)|} \sum_{h' \in g^{-1}(h)} \bar{q}_{h', i} \bar{k}_{h, i},
\end{aligned}
\label{eq:importance_score}
\end{equation}
where $g(h')$ maps query head $h'$ to its shared KV head.
For each KV head $h$, $\mathcal{S}_h$ contains the top-$r\%$ channel pairs by $\sigma_{h,i}$.
For a batch of size $B$, we store the selected pairs as a binary mask of size $B \times L \times H^{kv} \times (d/2)$ bits.

\subsection{Online KV-Cache Quantization Scheme}
\label{subsec:online-quant}

During decoding, PN-Mix keeps the first $N_s$ sink tokens and the residual buffer of size $N$ in FP16 for both keys and values.
When the buffer is full, the buffered keys are quantized with per-channel asymmetric groups: channel pairs in $\mathcal{S}_h$ use $b_H$ bits, and the remaining pairs use $b_L$ bits.
For 1-bit groups, we apply the correction in Eq.~\ref{eq:1bit_correction}~\citep{jie2025specache}.
This shifts the two reconstruction levels from the endpoints toward the bin midpoints, reducing the bias of extreme 1-bit min--max quantization.
Motivated by the lower quantization sensitivity of value caches observed in \S\ref{subsec:kv-sensitivity}, PN-Mix applies per-token asymmetric group quantization to values at $b_V$ bits.
\begin{equation}
s_{X} = \tfrac{\max X-\min X}{2}, \quad z_{X} = \tfrac{3\min X+\max X}{4}.
\label{eq:1bit_correction}
\end{equation}

\begin{table*}[t]
\centering
\small
\setlength{\tabcolsep}{4.5pt}
\renewcommand{\arraystretch}{0.95}
\begin{tabular}{llcccccc}
\toprule
Model & Method & K bits & GSM8K & MATH-Alg. & HumanEval & GPQA-Dia & Avg. \\
\midrule
\rowcolor{gray!15}
\cellcolor{white}\multirow{7}{*}{Llama3.1-8B-Instruct} & FP16 & 16 & $\mathrm{76.83}_{\pm 0.90}$ & $\mathrm{48.17}_{\pm 0.89}$ & $\mathrm{64.51}_{\pm 1.70}$ & $\mathrm{26.63}_{\pm 1.22}$ & $\mathrm{54.04}$ \\
 & KIVI        & 2.25 & $\mathrm{66.97}_{\pm 1.07}$ & $\mathrm{35.27}_{\pm 0.80}$ & $\mathrm{55.49}_{\pm 3.18}$ & $\mathrm{23.67}_{\pm 2.06}$ & $\mathrm{45.35}$ \\
 & PolarQuant  & 2.25 & $\mathrm{26.85}_{\pm 0.49}$ & $\mathrm{10.38}_{\pm 0.85}$ & $\mathrm{34.27}_{\pm 1.30}$ & $\mathrm{20.41}_{\pm 2.35}$ & $\mathrm{22.98}$ \\
 & ChanMix     & 2.32 & $\mathrm{70.51}_{\pm 0.75}$ & $\mathrm{37.57}_{\pm 0.70}$ & $\mathrm{59.63}_{\pm 2.23}$ & $\mathrm{23.37}_{\pm 1.38}$ & $\mathrm{47.77}$ \\
 & Kitty       & 2.39 & $\mathrm{74.51}_{\pm 0.62}$ & $\mathrm{46.25}_{\pm 0.72}$ & $\mathrm{62.68}_{\pm 1.78}$ & $\mathrm{23.98}_{\pm 2.14}$ & $\mathrm{51.86}$ \\
 & PN-Mix (1)  & 2.26 & $\mathbf{75.66}_{\pm 0.94}$ & $\mathbf{46.57}_{\pm 1.43}$ & $\mathbf{64.39}_{\pm 2.10}$ & $\mathrm{25.71}_{\pm 2.59}$ & $\mathbf{53.08}$ \\
 & PN-Mix (2)  & 2.29 & $\mathrm{75.65}_{\pm 0.61}$ & $\mathrm{46.28}_{\pm 1.24}$ & $\mathrm{63.78}_{\pm 2.30}$ & $\mathbf{26.33}_{\pm 3.02}$ & $\mathrm{53.01}$ \\
\midrule
\rowcolor{gray!15}
\cellcolor{white}\multirow{7}{*}{Qwen3-8B} & FP16 & 16 & $\mathrm{94.27}_{\pm 0.64}$ & $\mathrm{88.41}_{\pm 0.90}$ & $\mathrm{85.85}_{\pm 0.46}$ & $\mathrm{42.86}_{\pm 0.56}$ & $\mathrm{77.85}$ \\
 & KIVI        & 2.25 & $\mathrm{88.20}_{\pm 0.74}$ & $\mathrm{69.47}_{\pm 0.23}$ & $\mathrm{79.88}_{\pm 2.18}$ & $\mathrm{35.10}_{\pm 0.59}$ & $\mathrm{68.16}$ \\
 & PolarQuant  & 2.25 & $\mathrm{1.15}_{\pm 0.09}$ & $\mathrm{0.00}_{\pm 0.00}$ & $\mathrm{2.07}_{\pm 0.30}$ & $\mathrm{0.82}_{\pm 0.69}$ & $\mathrm{1.01}$ \\
 & ChanMix $\dagger$ & 2.32 & $\mathrm{90.34}_{\pm 5.86}$ & $\mathrm{86.57}_{\pm 4.01}$ & $\mathrm{83.65}_{\pm 1.30}$ & $\mathrm{38.97}_{\pm 0.76}$ & $\mathrm{74.88}$ \\
 & Kitty       & 2.39 & $\mathrm{91.93}_{\pm 0.44}$ & $\mathrm{82.29}_{\pm 0.68}$ & $\mathrm{83.90}_{\pm 1.62}$ & $\mathrm{39.59}_{\pm 2.53}$ & $\mathrm{74.43}$ \\
 & PN-Mix (1)  & 2.26 & $\mathrm{93.31}_{\pm 0.31}$ & $\mathbf{88.76}_{\pm 0.38}$ & $\mathbf{86.34}_{\pm 0.83}$ & $\mathrm{40.41}_{\pm 2.20}$ & $\mathbf{77.21}$ \\
 & PN-Mix (2)  & 2.29 & $\mathbf{93.89}_{\pm 0.25}$ & $\mathrm{87.40}_{\pm 0.46}$ & $\mathrm{84.87}_{\pm 2.26}$ & $\mathbf{41.84}_{\pm 2.21}$ & $\mathrm{76.66}$ \\
\midrule
\rowcolor{gray!15}
\cellcolor{white}\multirow{7}{*}{Qwen2.5-14B-Instruct} & FP16 & 16 & $\mathrm{90.83}_{\pm 0.29}$ & $\mathrm{78.16}_{\pm 0.29}$ & $\mathrm{77.07}_{\pm 2.46}$ & $\mathrm{40.00}_{\pm 3.27}$ & $\mathrm{71.52}$ \\
 & KIVI        & 2.25 & $\mathrm{85.66}_{\pm 1.07}$ & $\mathrm{72.00}_{\pm 1.26}$ & $\mathrm{71.71}_{\pm 2.43}$ & $\mathrm{36.73}_{\pm 0.72}$ & $\mathrm{66.52}$ \\
 & PolarQuant  & 2.25 & $\mathrm{61.47}_{\pm 1.18}$ & $\mathrm{44.53}_{\pm 1.04}$ & $\mathrm{58.17}_{\pm 1.71}$ & $\mathrm{27.76}_{\pm 2.77}$ & $\mathrm{47.98}$ \\
 & ChanMix     & 2.32 & $\mathrm{88.79}_{\pm 0.53}$ & $\mathrm{73.95}_{\pm 0.66}$ & $\mathrm{74.88}_{\pm 2.95}$ & $\mathbf{41.02}_{\pm 3.19}$ & $\mathrm{69.66}$ \\
 & Kitty       & 2.39 & $\mathbf{90.46}_{\pm 0.79}$ & $\mathbf{80.39}_{\pm 0.93}$ & $\mathrm{77.20}_{\pm 3.38}$ & $\mathrm{40.00}_{\pm 2.38}$ & $\mathrm{72.01}$ \\
 & PN-Mix (1)  & 2.26 & $\mathrm{90.07}_{\pm 0.55}$ & $\mathrm{79.58}_{\pm 0.87}$ & $\mathbf{78.17}_{\pm 2.13}$ & $\mathrm{40.61}_{\pm 2.51}$ & $\mathbf{72.11}$ \\
 & PN-Mix (2)  & 2.29 & $\mathrm{89.01}_{\pm 0.46}$ & $\mathrm{78.15}_{\pm 1.54}$ & $\mathrm{77.44}_{\pm 1.02}$ & $\mathrm{40.71}_{\pm 1.81}$ & $\mathrm{71.33}$ \\
\midrule
\rowcolor{gray!15}
\cellcolor{white}\multirow{7}{*}{Qwen3-32B} & FP16 & 16 & $\mathrm{90.80}_{\pm 0.42}$ & $\mathrm{84.62}_{\pm 1.00}$ & $\mathrm{85.61}_{\pm 1.13}$ & $\mathrm{53.27}_{\pm 2.20}$ & $\mathrm{78.58}$ \\
 & KIVI        & 2.25 & $\mathrm{87.47}_{\pm 0.62}$ & $\mathrm{59.87}_{\pm 1.15}$ & $\mathrm{84.51}_{\pm 0.62}$ & $\mathrm{46.73}_{\pm 1.53}$ & $\mathrm{69.65}$ \\
 & PolarQuant  & 2.25 & $\mathrm{63.21}_{\pm 0.64}$ & $\mathrm{1.31}_{\pm 0.13}$ & $\mathrm{46.71}_{\pm 2.27}$ & $\mathrm{7.55}_{\pm 1.63}$ & $\mathrm{29.70}$ \\
 & ChanMix $\dagger$ & 2.32 & $\mathrm{90.08}_{\pm 0.84}$ & $\mathrm{82.73}_{\pm 0.78}$ & $\mathrm{86.46}_{\pm 2.36}$ & $\mathrm{48.78}_{\pm 2.15}$ & $\mathrm{77.01}$ \\
 & Kitty       & 2.39 & $\mathbf{91.10}_{\pm 0.54}$ & $\mathrm{76.73}_{\pm 0.98}$ & $\mathrm{87.68}_{\pm 2.36}$ & $\mathrm{47.86}_{\pm 1.53}$ & $\mathrm{75.84}$ \\
 & PN-Mix (1)  & 2.26 & $\mathrm{90.33}_{\pm 0.30}$ & $\mathrm{82.81}_{\pm 0.60}$ & $\mathbf{88.29}_{\pm 1.36}$ & $\mathbf{50.82}_{\pm 1.39}$ & $\mathbf{78.06}$ \\
 & PN-Mix (2)  & 2.29 & $\mathrm{90.61}_{\pm 0.41}$ & $\mathbf{83.91}_{\pm 0.71}$ & $\mathrm{85.37}_{\pm 1.77}$ & $\mathrm{48.88}_{\pm 0.59}$ & $\mathrm{77.19}$ \\
 \bottomrule
\end{tabular}
\caption{\textbf{Accuracy comparison on standard reasoning and code-generation benchmarks.}
All baselines are configured to use an average key-cache precision in the 2.25--2.4-bit range, and the value cache is kept in FP16.
Each cell reports the mean $\pm$ standard deviation over five runs. 
Bold indicates the best non-FP16 result in each column.
$\dagger$ denotes the results obtained by applying the reproduced profiling procedure and the resulting profile to the model.}
\label{tab:normal_cot}
\end{table*}

\begin{table*}[t]
\centering
\small
\setlength{\tabcolsep}{5pt}
\renewcommand{\arraystretch}{0.95}
\begin{tabular}{llccccc}
\toprule
\multirow{2}{*}{Model} & \multirow{2}{*}{Method} & \multirow{2}{*}{K bits} & \multicolumn{2}{c}{AIME24} & \multicolumn{2}{c}{AIME25} \\
\cmidrule(lr){4-5} \cmidrule(lr){6-7}
 & & & $b_V{=}16$ & $b_V{=}2$ & $b_V{=}16$ & $b_V{=}2$ \\
\midrule
\rowcolor{gray!15}
\cellcolor{white}\multirow{7}{*}{R1-Llama-8B} & FP16 & 16 & \multicolumn{2}{>{\columncolor{gray!15}}c}{$\mathrm{48.33}_{\pm 6.19}$} & \multicolumn{2}{>{\columncolor{gray!15}}c}{$\mathrm{31.33}_{\pm 8.19}$} \\
 & KIVI        & 2.25 & $\mathrm{28.67}_{\pm 7.02}$ & $\mathrm{12.00}_{\pm 4.99}$ & $\mathrm{26.67}_{\pm 5.16}$ & $\mathrm{19.33}_{\pm 3.59}$ \\
 & PolarQuant  & 2.25 & $\mathrm{0.00}_{\pm 0.00}$ & $\mathrm{0.00}_{\pm 0.00}$ & $\mathrm{0.00}_{\pm 0.00}$ & $\mathrm{0.00}_{\pm 0.00}$ \\
 & ChanMix     & 2.32 & $\mathrm{28.00}_{\pm 6.70}$ & $\mathrm{19.00}_{\pm 5.59}$ & $\mathrm{26.00}_{\pm 3.27}$ & $\mathrm{16.67}_{\pm 4.47}$ \\
 & Kitty       & 2.39 & $\mathrm{42.00}_{\pm 5.62}$ & $\mathrm{42.00}_{\pm 5.42}$ & $\mathrm{35.00}_{\pm 6.19}$ & $\mathrm{30.00}_{\pm 5.96}$ \\
 & PN-Mix (1) & 2.26 & $\mathrm{48.33}_{\pm 5.43}$ & $\mathbf{47.67}_{\pm 6.16}$ & $\mathrm{29.67}_{\pm 7.22}$ & $\mathrm{31.00}_{\pm 4.23}$ \\
 & PN-Mix (2) & 2.29 & $\mathbf{50.67}_{\pm 5.54}$ &  $\mathrm{44.33}_{\pm 5.97}$  & $\mathbf{36.00}_{\pm 4.16}$ & $\mathbf{33.00}_{\pm 5.04}$ \\
\midrule
\rowcolor{gray!15}
\cellcolor{white}\multirow{7}{*}{Qwen3-8B} & FP16 & 16 & \multicolumn{2}{>{\columncolor{gray!15}}c}{$\mathrm{76.33}_{\pm 4.82}$} & \multicolumn{2}{>{\columncolor{gray!15}}c}{$\mathrm{67.00}_{\pm 4.58}$} \\
 & KIVI        & 2.25 & $\mathrm{64.00}_{\pm 5.54}$ & $\mathrm{56.33}_{\pm 4.33}$ & $\mathrm{55.33}_{\pm 4.99}$ & $\mathrm{51.33}_{\pm 3.06}$ \\
 & PolarQuant  & 2.25 & $\mathrm{0.00}_{\pm 0.00}$ & $\mathrm{0.00}_{\pm 0.00}$ & $\mathrm{0.00}_{\pm 0.00}$ & $\mathrm{0.00}_{\pm 0.00}$ \\
 & ChanMix$\dagger$     & 2.32 & $\mathrm{69.66}_{\pm 4.06}$ & $\mathrm{60.66}_{\pm 7.11}$ & $\mathrm{61.00}_{\pm 4.48}$ & $\mathrm{57.00}_{\pm 4.33}$ \\
 & Kitty       & 2.39 & $\mathrm{70.67}_{\pm 5.93}$ & $\mathrm{71.33}_{\pm 4.27}$ & $\mathrm{63.33}_{\pm 5.16}$ & $\mathrm{62.00}_{\pm 5.62}$ \\
 & PN-Mix (1) & 2.26 & $\mathrm{74.67}_{\pm 3.06}$ & $\mathbf{74.67}_{\pm 6.00}$ & $\mathrm{65.33}_{\pm 4.99}$ & $\mathrm{64.00}_{\pm 6.63}$ \\
 & PN-Mix (2) & 2.29 & $\mathbf{76.00}_{\pm 4.67}$ & $\mathrm{74.33}_{\pm 3.67}$ & $\mathbf{68.00}_{\pm 4.99}$ & $\mathbf{65.33}_{\pm 2.67}$ \\
\midrule
\rowcolor{gray!15}
\cellcolor{white}\multirow{7}{*}{R1-Qwen-7B} & FP16 & 16 & \multicolumn{2}{>{\columncolor{gray!15}}c}{$\mathrm{54.33}_{\pm 5.39}$} & \multicolumn{2}{>{\columncolor{gray!15}}c}{$\mathrm{39.33}_{\pm 7.57}$} \\
 & KIVI        & 2.25 & $\mathrm{35.33}_{\pm 4.76}$ & $\mathrm{25.00}_{\pm 7.64}$ & $\mathrm{25.00}_{\pm 4.53}$ & $\mathrm{21.00}_{\pm 4.23}$ \\
 & PolarQuant  & 2.25 & $\mathrm{8.67}_{\pm 4.52}$ & $\mathrm{9.00}_{\pm 3.96}$ & $\mathrm{9.67}_{\pm 3.48}$ & $\mathrm{7.00}_{\pm 2.77}$ \\
 & ChanMix$\dagger$     & 2.32 & $\mathrm{45.33}_{\pm 2.67}$ & $\mathrm{46.33}_{\pm 6.40}$ & $\mathbf{38.66}_{\pm 5.61}$ & $\mathrm{34.66}_{\pm 4.26}$ \\
 & Kitty       & 2.39 & $\mathrm{48.67}_{\pm 7.18}$ & $\mathrm{45.67}_{\pm 5.59}$ & $\mathrm{37.33}_{\pm 4.42}$ & $\mathbf{36.33}_{\pm 3.78}$ \\
 & PN-Mix (1) & 2.26 & $\mathbf{49.33}_{\pm 5.73}$ & $\mathbf{47.67}_{\pm 3.96}$ & $\mathrm{36.33}_{\pm 4.58}$ & $\mathrm{35.00}_{\pm 5.82}$ \\
 & PN-Mix (2) & 2.29 & $\mathrm{48.67}_{\pm 5.81}$ & $\mathrm{45.67}_{\pm 5.39}$ & $\mathrm{36.67}_{\pm 6.83}$ & $\mathrm{32.67}_{\pm 5.54}$ \\
\bottomrule
\end{tabular}
\caption{\textbf{Accuracy comparison on AIME.}
For each model, we report both settings where the value cache is kept in FP16 ($b_V{=}16$) and where it is quantized to the 2-bit range ($b_V{=}2$).
Each cell reports the mean $\pm$ standard deviation over 10 responses per question.
Bold indicates the best non-FP16 result in each column.
$\dagger$ denotes the results obtained by applying the reproduced profiling procedure and the resulting profile to the model.}
\label{tab:reasoning}
\end{table*}

\section{Experiments}
\label{sec:experiments}

\subsection{Experimental Setup}
\label{subsec:exp-setup}

\paragraph{Models and tasks.}
We evaluate accuracy with lm-evaluation-harness~\citep{eval-harness}.
For standard reasoning benchmarks, we use Llama-3.1-8B-Instruct~\citep{grattafiori2024llama3herdmodels}, Qwen3-8B/32B~\citep{yang2025qwen3technicalreport}, and Qwen2.5-14B-Instruct~\citep{qwen2025qwen25technicalreport} on GSM8K~\citep{cobbe2021gsm8k} (8-shot), MATH-Algebra~\citep{hendrycksmath2021} (4-shot), HumanEval~\citep{DBLP:journals/corr/abs-2107-03374}, and GPQA-Diamond~\citep{rein2024gpqa} (5-shot), with a 4096-token generation limit.
For AIME 2024~\citep{aime24} and AIME 2025~\citep{aime25}, we use DeepSeek-R1-Distill-Llama-8B~\citep{deepseekai2025deepseekr1incentivizingreasoningcapability}, Qwen3-8B, and DeepSeek-R1-Distill-Qwen-7B~\citep{deepseekai2025deepseekr1incentivizingreasoningcapability} with $(\text{temperature}, \text{top-}p, \text{top-}k)=(0.6,0.95,20)$ and a 32768-token limit.
We report mean and standard deviation over five runs for standard benchmarks and 10 responses per question for AIME; for HumanEval, we report pass@1 based on unit-test correctness, and all other tasks use extracted-answer accuracy.

\paragraph{Baselines.}
We compare PN-Mix with uniform-precision baselines, KIVI~\citep{kivi} and PolarQuant~\citep{wu2025polarquant}, and mixed-precision baselines, ChanMix~\citep{liao2026channelaware} and Kitty~\citep{xia2025kittyaccurateefficient2bit}.
Unless otherwise specified, all baselines use group size $G=128$ and residual buffer length $N=128$.

For models without official ChanMix profiles, we use reproduced profiles following the original profiling procedure; 
details of the reproduction and validation are provided in Appendix~\ref{appendix:bit-calc}.

\paragraph{PN-Mix configurations.}
PN-Mix uses $N=128$, $b_H=3$, $b_L=1$, and $N_s=32$, following the baseline residual length and Kitty's sink-token setting.
We report two similar-budget operating points: PN-Mix~(1) ($r=50\%$, $G=128$, 2.26 bits) and PN-Mix~(2) ($r=40\%$, $G=64$, 2.29 bits), comparing a larger high-bit ratio with a smaller group size.

\paragraph{Hardware and software.}
Accuracy experiments use NVIDIA RTX A6000 GPUs, and inference-efficiency experiments use a single NVIDIA A100-SXM4-40GB GPU.
All experiments use PyTorch 2.4.1, CUDA 12.1, and FlashAttention~\citep{dao2022flashattention} 2.7.4.post1.
\subsection{Downstream Task Accuracy}
\label{subsec:exp-accuracy}

We evaluate PN-Mix accuracy in two settings: (i) standard reasoning benchmarks on general-purpose models (Table~\ref{tab:normal_cot}) and (ii) AIME benchmarks on reasoning models (Table~\ref{tab:reasoning}).

All baselines are configured to use 2.25--2.4 average key-cache bits; for channel-level mixed-precision methods, we keep their original bit-width design and adjust only the high-precision channel ratio.
These tables are intended as comparisons under comparable low key-cache bit budgets rather than best-effort evaluations at each method's recommended configuration.
The exact average bit calculation and method-specific settings are provided in Appendix~\ref{appendix:bit-calc}.
To isolate key-cache quantization, Table~\ref{tab:normal_cot} and the $b_V{=}16$ columns of Table~\ref{tab:reasoning} keep values in full precision.
For the $b_V{=}2$ columns of Table~\ref{tab:reasoning}, we follow each method's original value-cache quantization design to evaluate full KV-cache compression.

\paragraph{Standard reasoning and code-generation benchmarks on general-purpose models.}
Table~\ref{tab:normal_cot} highlights the challenge of uniform low-bit quantization. 
Uniform 2-bit KIVI incurs a 5.0--9.7 point drop from FP16 in the Avg. column across all four models.
In contrast, PN-Mix~(1), using an average key-cache precision of only 2.26 bits, remains within one point of FP16 in Avg. across all four models. 
More importantly, despite using fewer bits than ChanMix and Kitty, PN-Mix~(1) achieves comparable or higher Avg. scores across the evaluated Llama and Qwen models spanning 8B to 32B parameters. 
At the task level, the differences from Kitty on GSM8K are relatively small, while PN-Mix~(1) tends to achieve comparable or better results on HumanEval and GPQA-Diamond. 
Between the two PN-Mix operating points, PN-Mix~(1) yields a higher Avg. score for all four models while using slightly fewer bits than PN-Mix~(2).

\paragraph{AIME benchmarks on reasoning models.}
Table~\ref{tab:reasoning} shows how effectively PN-Mix preserves AIME24 and AIME25 accuracy on reasoning models under a low key-cache bit budget. 
Despite using fewer key bits than ChanMix and Kitty, PN-Mix records the highest mean accuracy among the non-FP16 methods in five of the six model--benchmark settings, with R1-Qwen-7B on AIME25 being the exception. 
This pattern is also observed when the value cache is quantized to 2 bits.

\subsection{Ablation Study}
\label{subsec:exp-ablation}

We further validate PN-Mix with two analyses, keeping the value cache in full precision to isolate key-cache quantization effects.

\begin{table}[t]
\centering
\small
\setlength{\tabcolsep}{3pt}
\begin{tabular}{lcccc}
\toprule
Importance score & 1-bit & sink & AIME24 & HumanEval \\
\midrule
\rowcolor{gray!15}
FP16                                     & --           & --           & $\mathrm{76.33}_{\pm 4.82}$ & $\mathrm{85.85}_{\pm 0.46}$ \\
\rowcolor{red!10}
$\tilde{Q}\tilde{K}$ pair norm           & $\checkmark$ & $\checkmark$ & $\mathrm{76.00}_{\pm 4.67}$ & $\mathrm{84.87}_{\pm 2.26}$ \\
\midrule
$\tilde{Q}\tilde{K}$ pair norm           & $\checkmark$ & $\times$     & $72.00_{\pm 5.42}$ & $84.02_{\pm 2.54}$ \\
$\tilde{Q}\tilde{K}$ pair norm           & $\times$     & $\checkmark$ & $69.67_{\pm 6.90}$ & $80.36_{\pm 2.48}$ \\
\midrule
$\tilde{K}$ range                        & $\checkmark$ & $\checkmark$ & $70.67_{\pm 3.59}$ & $83.78_{\pm 2.48}$ \\
$\tilde{K}$ magnitude                    & $\checkmark$ & $\checkmark$ & $73.00_{\pm 5.04}$ & $84.39_{\pm 1.95}$ \\
$\tilde{K}$ pair norm                    & $\checkmark$ & $\checkmark$ & $74.67_{\pm 4.00}$ & $84.63_{\pm 2.01}$ \\
$\tilde{Q}\tilde{K}$ pair norm*           & $\checkmark$ & $\checkmark$ & $72.33_{\pm 5.39}$ & $82.31_{\pm 0.01}$ \\
\bottomrule
\end{tabular}
\caption{\textbf{Ablation of PN-Mix design choices on Qwen3-8B.}
All variants use PN-Mix~(2) with FP16 values; the highlighted row is the final configuration and (*) denotes the online variant.}
\label{tab:ablation}
\end{table}

\paragraph{Design component ablation.}
Table~\ref{tab:ablation} summarizes the component ablations.
Removing either sink-token preservation or the 1-bit correction degrades accuracy, with the latter causing a larger drop, indicating that both components are important for stable low-bit key-cache quantization.
Replacing the proposed $\tilde{Q}\tilde{K}$ pair norm with $\tilde{K}$-only criteria lowers mean accuracy, especially on AIME24, supporting the use of query-side information.
The $\tilde{Q}\tilde{K}$ pair norm$^{*}$ row tests online re-selection, which accumulates pair-norm statistics during decoding and reselects salient pairs for each quantization group.
This online variant not only underperforms PN-Mix but also requires storing and loading group-specific salient channel-pair indices, indicating that prefill-based salient channel-pair selection is more effective and practical under the evaluated conditions.

\begin{figure}[t]
\centering
\includegraphics[width=\linewidth]{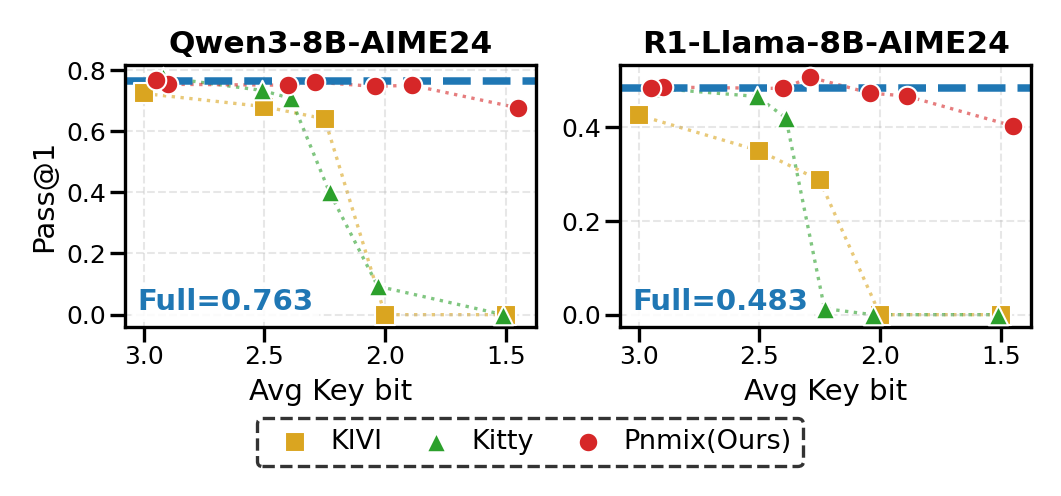}
\caption{\textbf{Pareto frontier in the 1.5--3.0-bit range}.
Detailed configurations and model accuracies are provided in Appendix~\ref{appendix:pareto-config}.}
\label{fig:pareto}
\end{figure}

\paragraph{Bit-rate Pareto frontier.}
While the main results focus near 2.25 average key-cache bits, we also evaluate AIME24 accuracy on Qwen3-8B and R1-Llama-8B at target bit-widths of 3.0, 2.5, 2.25, 2.0, and 1.5 bits.
For each method, we sweep applicable $b_H,b_L,G,r$ settings and select the closest configuration to each target; details are in Appendix~\ref{appendix:pareto-config}.
For Kitty, we keep its original $(b_H,b_L)=(4,2)$ design for ${\geq}2.25$ bits and additionally sweep $b_H,b_L$ below 2.25 bits.
Figure~\ref{fig:pareto} shows that although both Kitty and PN-Mix are close to FP16 at 2.5--3.0 bits, PN-Mix degrades more gracefully at lower bit-widths, indicating that its advantage extends beyond the 2.25-bit operating point.

\subsection{Inference Efficiency}
\label{subsec:exp-efficiency}

We compare PN-Mix with FP16 and Kitty in terms of memory scaling (Figure~\ref{appendix:peak-mem-fig}) and end-to-end latency (Table~\ref{tab:e2e_a100}) on Qwen3-8B using a single NVIDIA A100-SXM4-40GB GPU.
Kitty\footnote{\url{https://github.com/Summer-Summer/Kitty}} is evaluated using its official implementation with the original $(b_H,b_L)=(4,2)$, $r=25\%$ setting.
For PN-Mix, we use our Triton~\citep{tillet2019triton} kernel implementation (Appendix~\ref{appendix:kernel-implementation}) for the lower-bit PN-Mix~(1) setting, with $b_H=3$, $b_L=1$, and $b_V=2$.
In all efficiency experiments, each method preallocates KV-cache memory for \texttt{max\_cache\_len} tokens before running inference.

\paragraph{Memory scaling and throughput.}

We evaluate the maximum feasible generation length $L_{\mathrm{gen}}$ and batch size $B$ under the 40GB memory budget.

As shown in Figure~\ref{appendix:peak-mem-fig}, PN-Mix supports $L_{\mathrm{gen}}=32\mathrm{K}$ at $B=32$ and $B=110$ at $L_{\mathrm{gen}}=8\mathrm{K}$, which are 8$\times$ longer and 5.5$\times$ larger than FP16, respectively.
This larger feasible batch range also improves maximum throughput.
Appendix~\ref{appendix:peak-memory} shows that FP16 is faster at the same batch size when it fits in memory, whereas PN-Mix supports larger batches under the 40GB memory budget and reaches 2.37$\times$ the maximum per-GPU throughput achievable by FP16.

\begin{table}[t]
\centering
\footnotesize
\setlength{\tabcolsep}{3pt}
\begin{tabular}{clrrrr}
\toprule
\makecell{$B$/$P$/$L_{\mathrm{gen}}$} & Method & \makecell{Mem\\(GB)} & \makecell{TTFT\\(ms)} & \makecell{Decode\\(s)} & \makecell{Total\\(s)} \\
\midrule
\rowcolor{gray!15}
\cellcolor{white}\multirow{3}{*}{1/141/32K}
 & FP16   & 19.80 &   46 & 1299 & 1299 \\
 & Kitty  & 16.03 &   55 & 2121 & 2121 \\
 & PN-Mix & 15.95 &   71 & 2036 & 2036 \\
\midrule
\rowcolor{gray!15}
\cellcolor{white}\multirow{3}{*}{1/16K/8K}
 & FP16   & 20.26 & 1679 &  325 &  327 \\
 & Kitty  & 17.45 & 1702 &  590 &  592 \\
 & PN-Mix & 17.39 & 1714 &  563 &  565 \\
\midrule
\rowcolor{gray!15}
\cellcolor{white}\multirow{3}{*}{1/16K/32K}
 & FP16   & 23.63 & 1676 & 1302 & 1304 \\
 & Kitty  & 17.99 & 1706 & 3254 & 3256 \\
 & PN-Mix & 17.87 & 1715 & 3042 & 3044 \\
\midrule
\rowcolor{gray!15}
\cellcolor{white}\multirow{3}{*}{32/141/32K}
 & FP16   & \textit{OOM} &  --  &  --  &  --  \\
 & Kitty  & \textit{OOM} &  --  &  --  &  --  \\
 & PN-Mix & 37.12 &  396 & 3943 & 3943 \\
\bottomrule
\end{tabular}
\caption{\textbf{End-to-end latency and peak memory on a single A100-SXM4-40GB.}
$B$, $P$, and $L_{\mathrm{gen}}$ denote batch size, prompt length, and generation length.
$P{=}141$ corresponds to a single AIME24 problem, and $P{=}16\mathrm{K}$ to a GSM8K 88-shot prompt.}
\label{tab:e2e_a100}
\end{table}

\paragraph{End-to-end latency.}
Table~\ref{tab:e2e_a100} reports four representative $(B,P,L_{\mathrm{gen}})$ settings, testing prefill overhead under long prompts and decode-time savings relative to Kitty.
For $P=16\mathrm{K}$, PN-Mix differs from Kitty in TTFT by at most 12 ms, suggesting negligible prefill-stage prediction overhead.
Figure~\ref{appendix:per-step-fig} shows that PN-Mix reduces marginal per-step latency growth with KV-cache length by 8.8\% over Kitty on the long-CoT trajectory ($B=1$, $P=141$, $L_{\mathrm{gen}}=32\mathrm{K}$).
This translates into 4.0--6.5\% lower decode time across the three $B=1$ settings.
At $B=32$, $P=141$, $L_{\mathrm{gen}}=32\mathrm{K}$, only PN-Mix fits in memory, using 37.12 GB.

\section{Conclusion}

We presented PN-Mix, a pair-norm-based mixed-precision key-cache quantization method for RoPE-based decoder-only LLMs, designed to support efficient long-output inference.
PN-Mix allocates higher precision to RoPE channel pairs with large query--key pair-norm products and estimates these salient pairs once during prefill, avoiding offline calibration and repeated decoding-time saliency recomputation.
Across standard reasoning, code generation, and AIME benchmarks, PN-Mix achieves comparable or better average accuracy than prior mixed-precision baselines with lower average key-cache precision.
It also remains robust in the sub-2.25-bit regime and expands the feasible serving region for long-output inference.

\section*{Limitations}

PN-Mix is most directly applicable to RoPE-based decoder-only LLMs, as its saliency score is defined over RoPE channel pairs. Extending the same formulation to models without RoPE-style channel pairs may require additional adaptation.

Our efficiency results are measured on a single NVIDIA A100-SXM4-40GB GPU with our current Triton-based implementation. Since wall-clock latency depends on hardware, kernel optimization, batching strategy, and serving infrastructure, the exact speedup of PN-Mix may vary across deployment environments.

\section*{Acknowledgements}
This paper was a result of the research project supported by SK hynix Inc.

\bibliography{references}

\appendix


\section{Analysis on Key- and Value-Cache Quantization Sensitivity}
\label{appendix:kv-cache-sensitivity}

Table~\ref{appendix:kv-sink-compare} compares key-only and value-only KV cache quantization under the same average bit-width using KIVI~\citep{kivi}. 
Across both models and AIME benchmarks, value-only quantization is less harmful than key-only quantization, especially when the first 32 tokens are preserved in FP16. 
These results indicate that low-bit key-cache quantization is the main source of accuracy degradation, supporting our focus on mixed-precision key-cache quantization.

\section{Key-Cache Quantization Simulation Details}
\label{appendix:k-quant-simul}

To set up the simulation, we first collect representative inference traces from each model.
For R1-Qwen-7B and R1-Llama-8B, we randomly select one problem from AIME25 and LiveCodeBench~v5, respectively.
We run each model on the selected problem and collect the post-RoPE query and key states from all layers and heads throughout inference.

Table~\ref{appendix:k-quant-simul-table} provides the full results of the key-cache quantization simulation used in \S\ref{subsec:channel-pair-importance} and \S\ref{subsec:eff-importance}.
We report key-cache MSE $\Delta_{\tilde{K}}$ and attention KL divergence $D_{\mathrm{KL}}^{A}$ for uniform-precision baselines and H3(50\%)L1 mixed-precision settings.
Uniform baselines include per-channel asymmetric quantization following KIVI~\citep{kivi} and per-channel quantization after polar-coordinate conversion following PolarQuant~\citep{wu2025polarquant}.
For mixed precision, all settings use the same bit-width and differ only in the importance score and whether importance is computed from the full sequence or prefill tokens only.

The results show that $\tilde{Q}\tilde{K}$ pair norm consistently yields the lowest attention KL divergence among the mixed-precision criteria.
Importantly, the prefill-token-based version closely matches the full-sequence version, supporting our choice of estimating salient channel pairs once during prefill.

\section{Salient Channel-Pair Prediction Accuracy}
\label{appendix:ranking}

\begin{figure}[t]
\centering
\includegraphics[width=\linewidth]{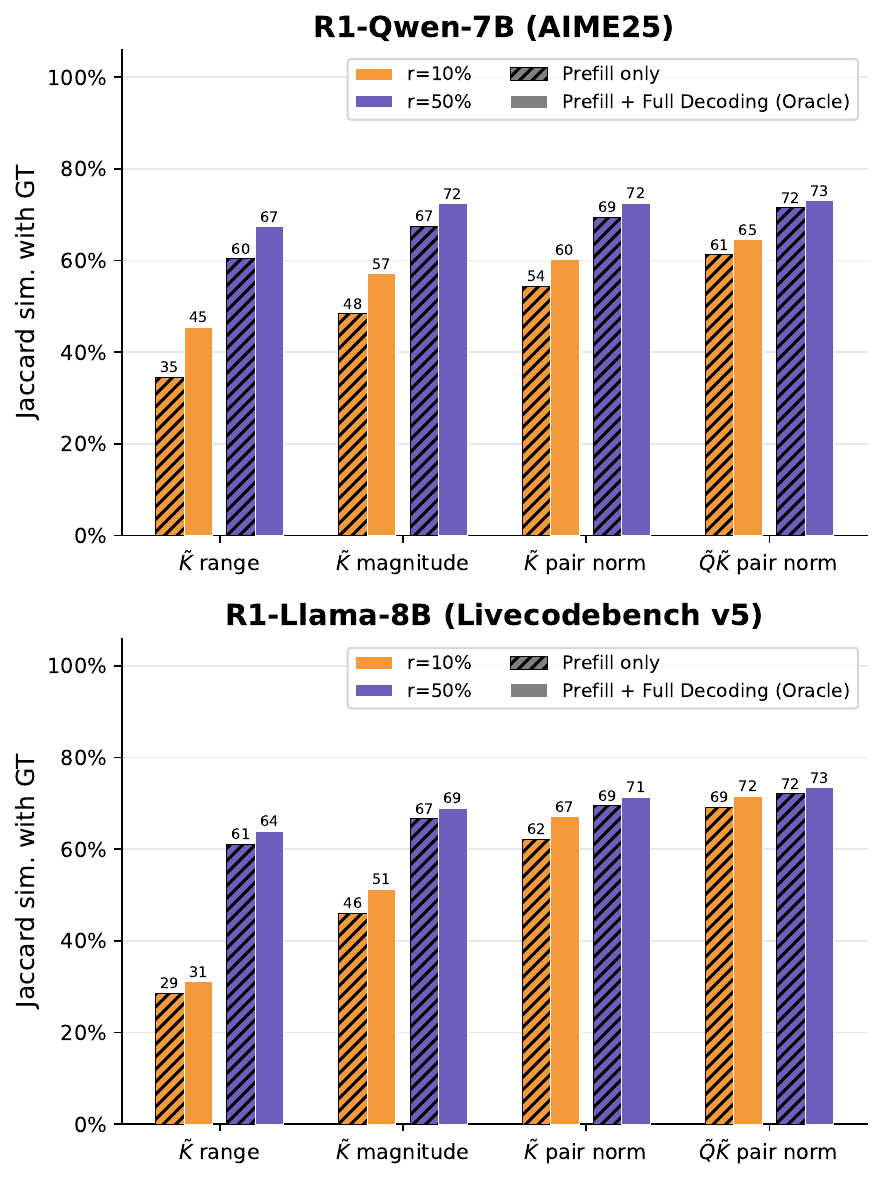}
\caption{\textbf{Salient channel-pair prediction accuracy.} Jaccard similarity between the top-$r\%$ channel-pair predictions of four importance criteria and the decoding-time ground truth, averaged over layers, heads, and decoding steps. Results are shown for $r=10\%$ and $r=50\%$.}
\label{appendix:ranking_appendix_bar}
\end{figure}

We assess how accurately different importance criteria identify salient RoPE channel pairs during decoding.
For each decoding-time query--key pair $(t,s)$, the ground-truth salient set is defined as the top-$r\%$ channel pairs ranked by $\|\mathbf{q}^{t}_{i}\|_2\|\mathbf{k}^{s}_{i}\|_2$.
Prediction accuracy is quantified by the Jaccard similarity between this set and the top-$r\%$ predictions obtained using four importance criteria: $\tilde{K}$ range, $\tilde{K}$ magnitude, $\tilde{K}$ pair norm, and $\tilde{Q}\tilde{K}$ pair norm.
For each criterion, two prediction settings are considered.
The first uses only prefill states to make the prediction, while the second uses the entire sequence, including prefill and all decoding states.
The Jaccard similarity of both predictions is computed against the decoding-time salient sets.

Figure~\ref{appendix:ranking_appendix_bar} shows the results on DeepSeek-R1-Distill-Qwen-7B and DeepSeek-R1-Distill-Llama-8B for $r\in\{10\%,50\%\}$.
Across both models and both values of $r$, the $\tilde{Q}\tilde{K}$ pair norm achieves the highest Jaccard similarity with the decoding-time ground truth, followed by $\tilde{K}$ pair norm, $\tilde{K}$ magnitude, and $\tilde{K}$ range.
The advantage of the $\tilde{Q}\tilde{K}$ pair norm is larger at $r=10\%$, indicating that incorporating query-side information is particularly useful for identifying the most salient channel pairs.
Moreover, for the $\tilde{Q}\tilde{K}$ pair norm, using only prefill states yields nearly the same Jaccard similarity as using the entire sequence.
This suggests that incorporating decoding states provides little additional benefit for salient-pair prediction.

\section{Concentration of Pair-Norm Importance Scores}
\label{appendix:sigma}

\begin{figure}[t]
\centering
\includegraphics[width=\linewidth]{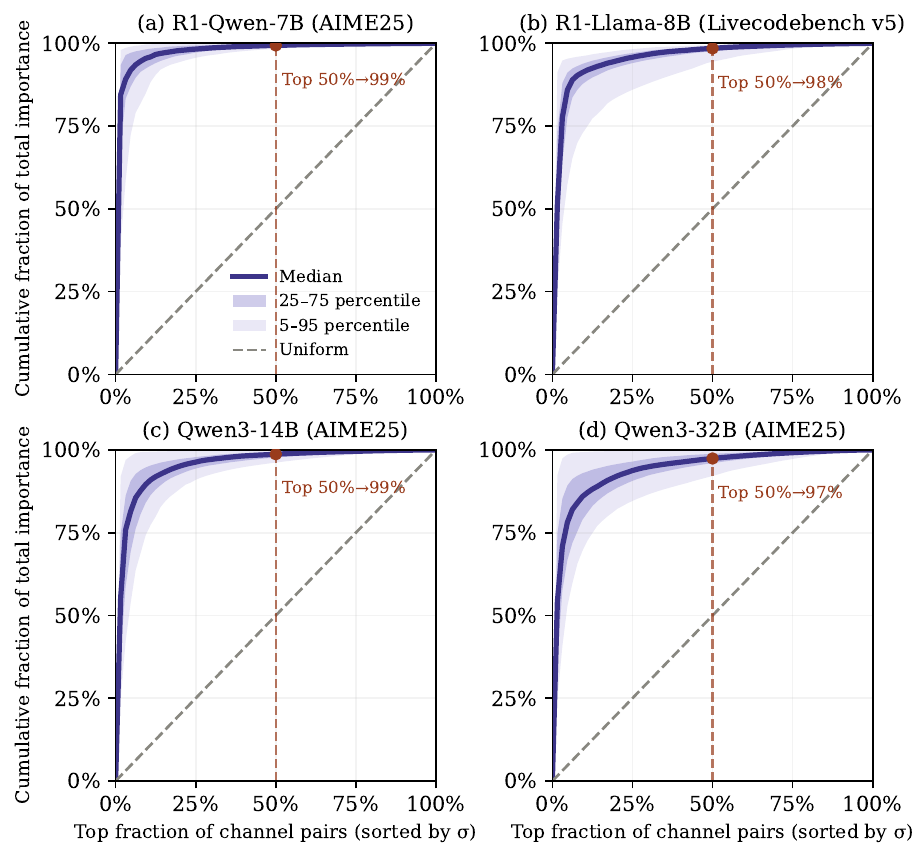}
\caption{\textbf{Cumulative ratio of channel-pair importance scores $\sigma_{h,i}$.}
The figure reports the cumulative importance ratio captured by the top $r\%$ channel pairs for R1-Qwen-7B, Qwen3-14B, and Qwen3-32B on AIME25, and for R1-Llama-8B on LiveCodeBench~v5.
For each layer and KV head, we compute the cumulative importance ratio from prefill-based scores and report the median of these ratios over layers, KV heads, and prompts.}
\label{appendix:sigma_fig}
\end{figure}

PN-Mix assigns higher precision to a subset of salient RoPE channel pairs identified by $\sigma_{h,i}$. Figure~\ref{appendix:sigma_fig} therefore reports the cumulative fraction of the total importance captured by the top-$r\%$ channel pairs. For each layer and KV head, we compute this cumulative fraction using importance scores obtained from prefill tokens and report the median across layers, KV heads, and prompts.

Across R1-Qwen-7B and R1-Llama-8B, the importance scores are highly concentrated in a relatively small subset of channel pairs. We further examine this concentration at larger model scales using Qwen3-14B and Qwen3-32B. On Qwen3-14B, the top 10\%, 30\%, and 50\% channel pairs account for 89.3\%, 97.0\%, and 98.7\% of the total importance, respectively. On Qwen3-32B, the corresponding ratios are 86.0\%, 94.8\%, and 97.4\%, showing that the importance distribution remains highly concentrated at 32B scale.

\section{Memory Scaling and Throughput Analysis}
\label{appendix:peak-memory}

\begin{figure}[!t]
\centering
\includegraphics[width=\linewidth]{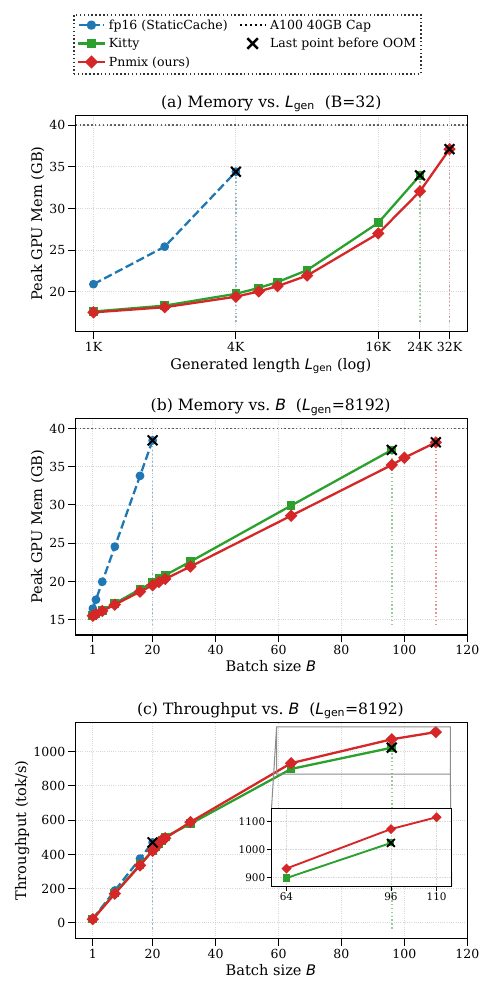}
\caption{\textbf{Memory scaling and generation throughput on a single A100-SXM4-40GB GPU.}
We compare FP16, Kitty, and PN-Mix on Qwen3-8B by sweeping
(a) generation length $L_{\mathrm{gen}}$ at fixed batch size $B=32$,
(b) batch size $B$ at fixed generation length $L_{\mathrm{gen}}=8\mathrm{K}$,
and (c) token generation throughput over $B$ at $L_{\mathrm{gen}}=8\mathrm{K}$.}
\label{appendix:peak-mem-fig}
\end{figure}

Figure~\ref{appendix:peak-mem-fig} evaluates how peak GPU memory and generation throughput scale with generation length and batch size on Qwen3-8B using a single NVIDIA A100-SXM4-40GB GPU. We compare FP16, Kitty, and PN-Mix. For all methods, the KV cache is preallocated up to \texttt{max\_cache\_len} before inference, allowing peak memory to be compared under the same static cache-allocation protocol.

Panel~(a) shows peak GPU memory as the generation length $L_{\mathrm{gen}}$ increases at a fixed batch size of $B=32$. The memory usage of FP16 grows rapidly with generation length and exceeds the 40GB GPU memory budget beyond $L_{\mathrm{gen}}=4\mathrm{K}$. Both Kitty and PN-Mix reduce this growth through KV-cache quantization, with PN-Mix showing the lowest memory usage among the three methods. PN-Mix remains within the memory budget up to $L_{\mathrm{gen}}=32\mathrm{K}$, enabling an 8$\times$ longer generation than FP16 at the same batch size.

Panel~(b) considers the complementary setting, fixing $L_{\mathrm{gen}}=8\mathrm{K}$ while increasing the batch size. Both quantized methods support substantially larger batches than FP16, with PN-Mix maintaining the lowest memory usage as $B$ increases. FP16 supports at most $B=20$ within the 40GB memory budget, whereas PN-Mix scales to $B=110$. Thus, PN-Mix supports a 5.5$\times$ larger batch size than FP16 at the same generation length. Together, panels~(a) and~(b) show that reducing the KV-cache footprint expands the feasible operating region along both the generation-length and batch-size dimensions.

Lower memory usage, however, does not necessarily imply higher throughput at the same batch size. Panel~(c) therefore reports per-GPU generation throughput as the batch size is increased at $L_{\mathrm{gen}}=8\mathrm{K}$, using a prompt length of $P=141$ and greedy decoding.

When the FP16 KV cache fits in memory, FP16 remains faster at the same batch size. For example, at $B=20$, FP16 achieves 469.8 tok/s compared with 420.8 tok/s for PN-Mix. However, FP16 already consumes 38.48 GB at this point and runs out of memory at $B=21$, limiting its maximum feasible throughput to 469.8 tok/s.

PN-Mix can continue increasing the batch size because of its smaller KV-cache footprint. It exceeds the maximum feasible FP16 throughput at $B=23$, reaching 485.0 tok/s, and scales to $B=110$, where it achieves 1114.0 tok/s while using 38.21 GB of GPU memory. This corresponds to 2.37$\times$ the maximum per-GPU throughput achievable by FP16 under the same 40GB memory budget.

These results distinguish two deployment regimes. When GPU memory is sufficient for the desired batch size and generation length, FP16 provides higher same-batch throughput because it avoids quantization and dequantization overhead. In memory-bound settings with larger batches or longer generations, PN-Mix can run configurations that exceed the FP16 memory limit. This larger feasible batch size allows PN-Mix to achieve a higher maximum per-GPU throughput under the same memory budget.

\section{Per-Step Decoding Latency Analysis}
\label{appendix:per-step-efficiency}

\begin{figure}[t]
\centering
\includegraphics[width=\linewidth]{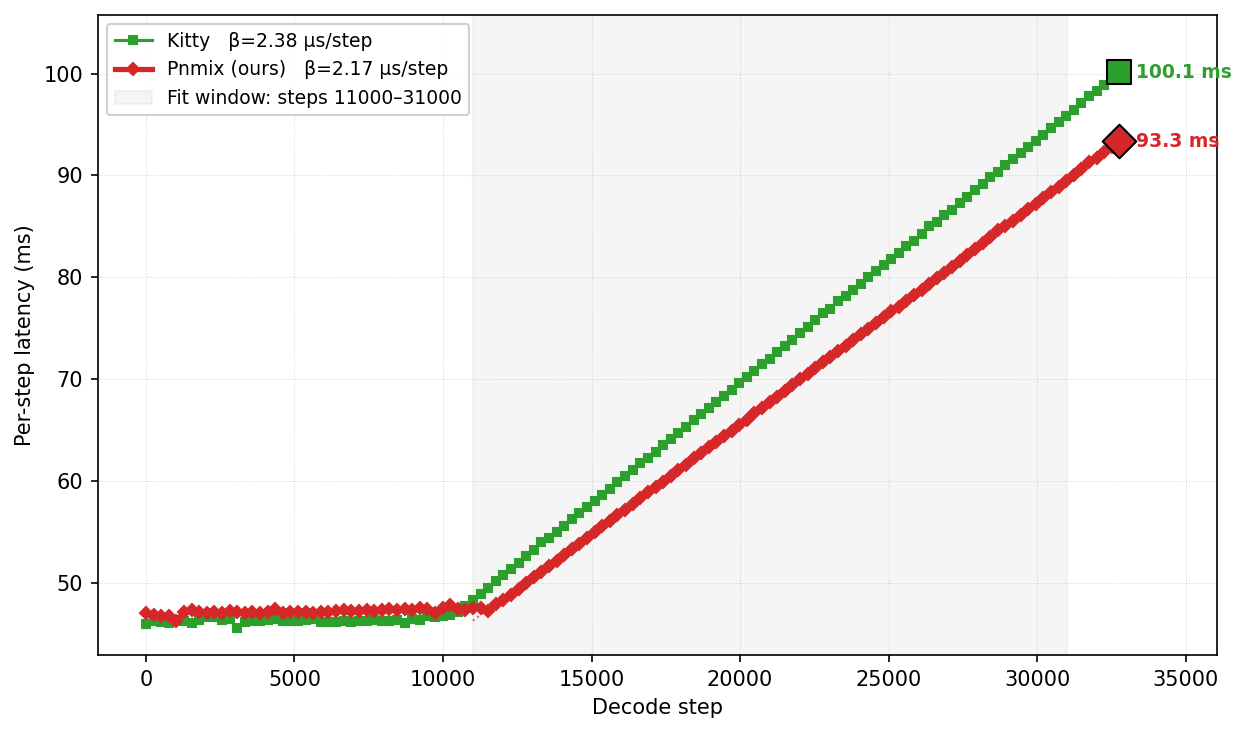}
\caption{\textbf{Per-step decode latency on a single A100-SXM4-40GB GPU.}
We measure Qwen3-8B decode latency with Kitty and PN-Mix.
The shaded region (steps 11K--31K) indicates the KV-cache-read-dominated regime, and $\beta$ denotes the linear-regression slope in this region ($\mu$s/step).}
\label{appendix:per-step-fig}
\end{figure}

Figure~\ref{appendix:per-step-fig} compares the per-step decoding latency of PN-Mix and Kitty along a long-CoT trajectory with $B=1$, $P=141$, and $L_{\mathrm{gen}}=32\mathrm{K}$, measured using \texttt{torch.cuda.Event}.
Both methods remain at a roughly 45--48 ms plateau up to approximately 10K decoding steps.
Beyond this point, per-step latency increases steadily with the growing KV-cache length.
Over the linear region from 11K to 31K steps, PN-Mix exhibits a lower latency-growth slope than Kitty (2.17 vs.\ 2.38 $\mu$s/step), corresponding to an 8.8\% reduction.
At 32K decoding steps, PN-Mix is also 6.8 ms faster than Kitty.
These results show that PN-Mix's lower average key-cache bit-width and absence of page-wise salient-channel index I/O lead to practical decoding latency improvements.

\section{Kernel Implementation}
\label{appendix:kernel-implementation}

We implement all kernels in Triton, including quantization and bit packing of the key and value caches, as well as the fused dequantization and attention operations during decoding.

The kernels are implemented for the configuration used in our downstream task evaluation, namely $b_H=3$, $b_L=1$, and $b_V=2$.
After mixed-precision quantization of the key cache, salient channel pairs are represented by integers in the range $[0,7]$, while the remaining channels are represented by integers in the range $[0,1]$.
Since the least significant 1 bit is shared by all channels and the upper 2 bits are meaningful only for salient channel pairs, our implementation packs these two parts into separate \texttt{uint8} tensors.
Specifically, we first extract the LSBs of all channels and pack every 8 consecutive token values of each channel into one byte, producing \texttt{packed\_low}, which is stored for all channels.
We then extract the upper 2 bits and pack every 4 consecutive token values of each channel into one byte, producing \texttt{packed\_hi}, which is stored only for the salient channel pairs.
For the value cache, after quantization, every 4 consecutive channel values of each token are packed into one byte and stored as a \texttt{uint8} tensor.

During the decoding-stage attention computation, we launch two Triton kernels: one for computing $\tilde{Q}_{h,t}\tilde{K}_{h,1:t}^{\top}/\sqrt{d}$ and the other for computing $A_{h,t}V_{h,1:t}$.
Each kernel sequentially processes the sink-token region, the quantized-cache region, and the residual-buffer region.
For the quantized region, the dequantized key or value cache is never written back to global memory.
Instead, each quantization group is loaded into on-chip registers, dequantized, and immediately consumed by the dot-product computation.

\section{Average Key Bit-Width Calculation and Method Configurations}
\label{appendix:bit-calc}

To compare the methods under a similar key-cache memory budget, we include both the quantized key values and the additional metadata required by each method in the bit-width calculation. Since the methods use different native precision-allocation schemes, Tables~\ref{tab:normal_cot} and~\ref{tab:reasoning} retain these schemes and use configurations within a narrow 2.25--2.4-bit range rather than enforcing an exact bit match. The resulting comparison therefore focuses on accuracy preservation in this low-bit regime rather than on the best-performing or recommended operating point of each method.

For a consistent calculation across methods, we exclude the FP16 residual buffer from both the storage cost and the number of key elements used for normalization. Let $T$ denote the total number of tokens, $N$ the FP16 residual-buffer length, $N_s$ the number of sink tokens retained in FP16, and
\begin{equation}
    T_q = T - N_s - N
\end{equation}
the number of quantized tokens. We further denote the token-axis quantization group size by $G$ and the head dimension by $d$. The effective key-storage cost includes the quantized key payload, FP16 sink tokens when applicable, and scale/zero-point storage. Saliency-related indices or masks are included separately as metadata overhead, and the sum of the two is reported as the total average key-cache bit-width. For sequence-length-dependent terms, we use $T_q=32{,}768$.

\paragraph{Uniform-precision methods.}
KIVI and PolarQuant use a uniform key bit-width $b$ and do not require saliency metadata. For each quantized key element, the payload contributes $b$ bits. Since each channel stores one FP16 scale and one FP16 zero-point for every group of $G$ tokens, the scale/zero-point overhead is $32/G$ bits per key element. Their effective key-storage cost is therefore
\begin{equation}
    \bar{b}_{K,\mathrm{storage}}^{\mathrm{KIVI/PolarQuant}}
    =
    b+\frac{32}{G}.
\end{equation}

\paragraph{Kitty and PN-Mix.}
Both Kitty and PN-Mix preserve $N_s$ sink tokens in FP16 and use mixed-precision key quantization with high and low precisions $b_H$ and $b_L$. For both methods, the effective key-storage cost consists of three components: the FP16 sink tokens, the mixed-precision quantized key values, and the FP16 scale/zero-point values stored for each token group. We define the average storage cost per quantized key element as
\begin{equation}
    b_{\mathrm{mix}}
    =
    r_{\mathrm{eff}}b_H
    +(1-r_{\mathrm{eff}})b_L
    +\frac{32}{G}.
\end{equation}
Here, the first two terms represent the mixed-precision key payload, while $32/G$ accounts for the FP16 scale and zero-point stored for every group of $G$ tokens. Normalizing the total storage by the number of key elements excluding the residual buffer gives
\begin{equation}
    \bar{b}_{K,\mathrm{storage}}
    =
    \frac{16N_s + T_q b_{\mathrm{mix}}}
         {N_s + T_q}.
\end{equation}

The two methods differ in how the high-precision channels are selected and how the corresponding saliency information is stored. For Kitty, high precision is assigned at the channel level. Because the number of selected channels must be an integer, the effective high-precision ratio is
\begin{equation}
    r_{\mathrm{eff}}^{\mathrm{Kitty}}
    =
    \frac{\lfloor dr\rfloor}{d},
\end{equation}
where $r$ is the nominal selection ratio. Kitty stores a channel-selection index for every quantization page. Since each page stores one \texttt{uint8} index per channel, the metadata cost is distributed over the $G$ tokens in the page, resulting in
\begin{equation}
    \bar{b}_{K,\mathrm{meta}}^{\mathrm{Kitty}}
    =
    \frac{8T_q}{G(N_s+T_q)}.
\end{equation}

For PN-Mix, high precision is assigned at the RoPE-pair level, so the selected number of channels must be an even integer. The corresponding effective high-precision ratio is
\begin{equation}
    r_{\mathrm{eff}}^{\mathrm{PN\text{-}Mix}}
    =
    \frac{2\lfloor rd/2\rfloor}{d}.
\end{equation}
Unlike Kitty, PN-Mix determines the salient RoPE pairs once during prefill and shares the resulting mask across all quantization groups. Since there are $d/2$ RoPE pairs and each pair requires one bit to record whether it is assigned high precision, the total mask size is $d/2$ bits per KV head. Dividing this one-time mask cost by the number of key elements excluding the residual buffer gives the average metadata overhead
\begin{equation}
    \bar{b}_{K,\mathrm{meta}}^{\mathrm{PN\text{-}Mix}}
    =
    \frac{1}{2(N_s+T_q)},
\end{equation}
which becomes negligible for long sequences.

\paragraph{ChanMix.}
ChanMix starts from a 2-bit allocation for all channels and promotes selected retrieval KV heads to 4 bits. Among the remaining channels, the same number of channels are reassigned to 1 and 3 bits, so their average precision remains 2 bits. If $r_{\mathrm{eff}}^{\mathrm{CM}}$ denotes the actual fraction of promoted KV heads after integer rounding, the average payload is therefore
\begin{equation}
    4r_{\mathrm{eff}}^{\mathrm{CM}}
    +2\left(1-r_{\mathrm{eff}}^{\mathrm{CM}}\right)
    =
    2\left(1+r_{\mathrm{eff}}^{\mathrm{CM}}\right).
\end{equation}
The effective promoted-head ratio is
\begin{equation}
    r_{\mathrm{eff}}^{\mathrm{CM}}
    =
    \frac{\lfloor L H^{kv} r_h\rfloor}{L H^{kv}},
\end{equation}
where $r_h$ is the nominal retrieval-head ratio, $L$ is the number of layers, and $H^{kv}$ is the number of KV heads per layer. ChanMix stores its scale and zero-point values in FP8, so each token group contributes 16 bits per channel for these two values, or $16/G$ bits per key element. Its effective key-storage cost is thus
\begin{equation}
    \bar{b}_{K,\mathrm{storage}}^{\mathrm{ChanMix}}
    =
    2\left(1+r_{\mathrm{eff}}^{\mathrm{CM}}\right)
    +\frac{16}{G}.
\end{equation}

ChanMix additionally stores, for each channel, its assigned bit-width and an index that maps the quantized channel ordering back to the original channel ordering. These two metadata tensors are stored once and shared across all tokens. Their average metadata overhead is
\begin{equation}
    \bar{b}_{K,\mathrm{meta}}^{\mathrm{ChanMix}}
    =
    \frac{64}{N_s+T_q},
\end{equation}
which is negligible at the sequence lengths considered here.

Table~\ref{tab:bitwidth_config_summary} summarizes the exact configurations used in our experiments and the resulting bit-width calculation. Panel~(a) reports the method-specific quantization and selection parameters, including the effective selection ratios after integer rounding. Panel~(b) separately reports the effective key-storage cost and saliency-metadata overhead, whose sum gives the total average key-cache bit-width. The resulting configurations span approximately 2.25--2.39 average key bits.

For PolarQuant, the 2-bit radius/angle configuration is used specifically to place the method in this low-bit comparison regime. PolarQuant originally uses 3-bit per-channel quantization, and reducing both radius and angle channels to 2 bits can amplify quantization error after the polar-coordinate transformation. We therefore do not interpret the results in Tables~\ref{tab:normal_cot} and~\ref{tab:reasoning} as comparisons against each method's recommended operating point; rather, they evaluate accuracy preservation under comparable low key-cache budgets.

For ChanMix, official profiles are unavailable for Qwen3-8B/32B and DeepSeek-R1-Distill-Qwen-7B.
We therefore reproduce the profiling procedure from the original ChanMix method and validate it on three models with official profiles.
The reproduced profiles show Pearson correlations of 0.98--0.99 with the official profiles for both key-range and retrieval-head scores, with 90--91\% overlap in the selected top-30\% retrieval heads.
We then use the validated procedure for the three models without official profiles.
Results based on these reproduced profiles are marked with $\dagger$ in Tables~\ref{tab:normal_cot} and~\ref{tab:reasoning}.

\section{Configurations for the Bit-Rate Pareto Frontier}
\label{appendix:pareto-config}

Table~\ref{appendix:pareto-config-table} reports the detailed hyperparameter configurations and AIME24 accuracies for the operating points shown in Figure~\ref{fig:pareto}.
For each method, we sweep candidate configurations and select the operating point whose average key-cache bit-width is closest to each target bit budget: 3.0, 2.5, 2.25, 2.0, and 1.5 bits.
For PN-Mix, we sweep $b_H,b_L\in\{4,3,2,1\}$, $G\in\{64,128\}$, and $r\in\{0.1,0.2,\ldots,0.9\}$.
For KIVI, we sweep $b\in\{2,1\}$ and $G\in\{32,64,128\}$.

For Kitty, we follow its original design in the $\geq$2.25-bit regime by fixing $(b_H,b_L)=(4,2)$ and sweeping only $G$ and $r$.
In the sub-2.25-bit regime, where the original $(4,2)$ setting cannot reach the target bit budgets, we additionally sweep $(b_H,b_L)$ using the same candidate set as PN-Mix.
For both Kitty and PN-Mix, the first $N_s=32$ sink tokens are preserved in full precision, while the value cache is kept at $b_V=16$ for all operating points.

\section{Additional Evaluation on Long-Input and Multilingual Tasks}
\label{appendix:long-input-multilingual}

To evaluate whether PN-Mix generalizes beyond the reasoning-oriented long-output settings in the main experiments, we further evaluate it on long-input and multilingual tasks using Llama-3.1-8B-Instruct. We compare PN-Mix with KIVI, ChanMix, and Kitty under the same low-bit regime: all compressed methods use an average key-cache precision of 2.25--2.39 bits, while the value cache is kept in full precision.

\paragraph{RULER.}
Table~\ref{appendix:ruler} reports results on all 13 RULER~\citep{hsieh2024ruler} tasks at context lengths from 16K to 64K. PN-Mix achieves the highest average accuracy among the compressed methods at all context lengths and remains close to FP16 even at 64K. The advantage is particularly pronounced on the challenging multi-key retrieval task (MK3), where PN-Mix maintains nearly unchanged accuracy as the context grows, while the competing quantized methods degrade substantially.

\paragraph{InfiniteBench.}
Table~\ref{appendix:infinitebench} presents results on InfiniteBench~\citep{zhang-etal-2024-bench}, covering diverse long-input tasks including retrieval, dialogue QA, long-document QA, and summarization. PN-Mix (2) achieves the highest average score among the compressed methods (33.98), close to FP16 (34.15), demonstrating that its robustness extends beyond controlled retrieval tasks to diverse long-context workloads.

\paragraph{MGSM.}
We assess multilingual performance on MGSM~\citep{shi2023language} across the compared methods (Table~\ref{appendix:mgsm}). PN-Mix remains competitive across all evaluated languages, with PN-Mix (2) achieving the best compressed-model accuracy on Chinese, French, and Japanese, showing that its effectiveness is not limited to English generation.

Taken together, these results indicate that PN-Mix maintains a favorable accuracy--precision trade-off across diverse long-context and multilingual workloads.

\begin{table}[t]
\centering
\small
\setlength{\tabcolsep}{3pt}
\renewcommand{\arraystretch}{0.95}
\begin{tabular}{lccccc}
\toprule
Method & en & zh & fr & ja & bn \\
\midrule
\rowcolor{gray!15}
FP16 & $\mathrm{82.80}$ & $\mathrm{62.80}$ & $\mathrm{64.00}$ & $\mathrm{50.80}$ & $\mathrm{58.00}$ \\
KIVI        & $\mathrm{73.20}$ & $\mathrm{56.00}$ & $\mathrm{58.80}$ & $\mathrm{47.20}$ & $\mathrm{44.40}$ \\
ChanMix     & $\mathrm{74.80}$ & $\mathrm{54.40}$ & $\mathrm{55.20}$ & $\mathrm{45.60}$ & $\mathrm{50.40}$ \\
Kitty       & $\mathbf{80.40}$ & $\mathrm{57.20}$ & $\mathrm{62.40}$ & $\mathrm{48.80}$ & $\mathbf{55.20}$ \\
PN-Mix (1)  & $\mathbf{80.40}$ & $\mathrm{57.20}$ & $\mathrm{63.60}$ & $\mathrm{48.40}$ & $\mathrm{52.40}$ \\
PN-Mix (2)  & $\mathrm{80.00}$ & $\mathbf{59.20}$ & $\mathbf{65.20}$ & $\mathbf{53.60}$ & $\mathrm{52.00}$ \\
\bottomrule
\end{tabular}
\caption{\textbf{Accuracy comparison on MGSM with Llama-3.1-8B-Instruct.}
We evaluate the full 250-example test set using 8-shot native-language CoT. 
Performance is measured using flexible-extract accuracy.
All compressed configurations (i.e., every method except FP16) are matched to
an average key-cache precision of 2.25--2.39 bits, while the value cache is kept in full precision.
Bold indicates the best non-FP16 result in each column.}
\label{appendix:mgsm}
\end{table}

\section{Additional Comparison with PM-KVQ}
\label{appendix:pmkvq}

We include a direct comparison with PM-KVQ~\citep{liu2026pmkvq}, a layer-level mixed-precision KV-cache quantization method for long-CoT LLMs. PM-KVQ profiles the quantization sensitivity of each transformer block on calibration data and uses integer programming to assign a final bit-width (Fbit) to each block under a given memory budget. During inference, it progressively reduces the KV-cache precision from 16 bits toward the assigned Fbit as the block-level memory budget is filled. In contrast, PN-Mix allocates precision within each attention head at RoPE channel-pair granularity using input-adaptive saliency estimated during prefill, without offline calibration.

Following the comparison protocol used in our main experiments, we evaluate the methods within the 2.25--2.4 average-bit range under the $b_V=2$ setting. PM-KVQ uses layer-level 4/2-bit Fbit allocation, which admits only discrete operating points. Assigning a 2-bit Fbit to every layer removes cross-layer precision differentiation, so we instead use the lowest configurations that retain PM-KVQ's mixed-precision allocation. Specifically, PM-KVQ assigns a 4-bit Fbit to 2 of 32 layers for R1-Llama-8B and to 1 of 28 layers for R1-Qwen-7B, resulting in average KV bit-widths of approximately 2.38 and 2.32 bits, respectively. These configurations still give PM-KVQ a larger average KV-bit budget than PN-Mix, which operates at 2.26--2.28 bits.

As shown in Table~\ref{tab:pmkvq}, PN-Mix achieves comparable accuracy to PM-KVQ on R1-Llama-8B despite using the smaller average KV-bit budget. On R1-Qwen-7B, both PN-Mix variants achieve higher mean accuracy than PM-KVQ on AIME24 and AIME25 while using fewer average KV bits.

\begin{table}[t]
\centering
\small
\setlength{\tabcolsep}{3pt}
\renewcommand{\arraystretch}{0.95}

\begin{tabular}{lccc}
\toprule
\textbf{Method}
& \makecell{\textbf{Avg. KV}\\\textbf{bits}}
& \textbf{AIME24}
& \textbf{AIME25} \\
\midrule

\multicolumn{4}{l}{\textit{R1-Llama-8B}} \\
\rowcolor{gray!15}
\quad FP16       & 16   & $\mathrm{48.33}_{\pm 6.19}$ & $\mathrm{31.33}_{\pm 8.19}$ \\
\quad PM-KVQ     & 2.38 & $\mathrm{46.33}_{\pm 6.04}$ & $\mathrm{31.66}_{\pm 4.53}$ \\
\quad PN-Mix (1) & 2.26 & $\mathrm{47.67}_{\pm 6.16}$ & $\mathrm{31.00}_{\pm 4.23}$ \\
\quad PN-Mix (2) & 2.28 & $\mathrm{44.33}_{\pm 5.97}$ & $\mathrm{33.00}_{\pm 5.04}$ \\

\midrule

\multicolumn{4}{l}{\textit{R1-Qwen-7B}} \\
\rowcolor{gray!15}
\quad FP16       & 16   & $\mathrm{54.33}_{\pm 5.39}$ & $\mathrm{39.33}_{\pm 7.57}$ \\
\quad PM-KVQ     & 2.32   & $\mathrm{38.33}_{\pm 4.53}$ & $\mathrm{31.67}_{\pm 4.99}$ \\
\quad PN-Mix (1) & 2.26 & $\mathrm{47.67}_{\pm 3.96}$ & $\mathrm{35.00}_{\pm 5.82}$ \\
\quad PN-Mix (2) & 2.28 & $\mathrm{45.67}_{\pm 5.39}$ & $\mathrm{32.67}_{\pm 5.54}$ \\

\bottomrule
\end{tabular}
\caption{\textbf{Comparison with PM-KVQ on AIME24 and AIME25.}
For PN-Mix, we use the same $b_V=2$ configurations as in Table~\ref{tab:reasoning}, while PM-KVQ uses the lowest model-specific mixed-precision operating point within the 2.25--2.4 average-bit range. Each cell reports the mean $\pm$ standard deviation over 10 responses per question.}
\label{tab:pmkvq}
\end{table}

\section{Step-Wise Stability of Prefill-Based Saliency Selection}
\label{appendix:prefill_drift_decode}

Appendix~\ref{appendix:ranking} shows that channel pairs selected from the prefill states exhibit high overlap with decoding-time salient channel pairs on average. However, this averaged overlap does not reveal whether the overlap between the channel pairs selected during prefill and decoding-time salient channel pairs remains stable throughout generation or decreases at later decoding steps. We therefore examine how this overlap changes over the course of decoding under long-generation and multi-topic stress-test settings.

For each causal query--key pair $(t,s)$ at decoding step $t$, we compare the top-$r\%$ channel pairs selected from the prefill states with the top-$r\%$ salient channel pairs identified for that query--key pair. To prevent the overlap from being dominated by prefill keys, we restrict $s$ to keys generated during decoding up to step $t$. We use two reference scores to identify the decoding-time salient channel pairs: the pair-norm score, $\|\mathbf{q}_i^t\|_2\|\mathbf{k}_i^s\|_2$, and the angle-aware absolute pair dot product, $|\langle \mathbf{q}_i^t,\mathbf{k}_i^s\rangle|$. For reporting, we group the decoding steps into four consecutive ranges and average the overlap over all eligible query--key pairs, heads, and layers within each range. We additionally report the difference between the maximum and minimum overlap across 1K-step bins for the long-generation setting and across the four reporting bins for the multi-topic setting to summarize temporal variation.

Table~\ref{tab:prefill_drift_decode} shows that the overlap between the channel pairs selected during prefill and decoding-time salient channel pairs remains stable throughout decoding in both stress-test settings. In the long-generation setting, where R1-Qwen-7B generates 32,672 tokens from only 96 prefill tokens, the overlap for the $\tilde{Q}\tilde{K}$ pair-norm criterion at $r=50\%$ stays at approximately $0.72$ across all four decoding ranges under the pair-norm score and at approximately $0.68$ under the absolute pair dot product. This indicates that the overlap does not progressively decrease over the long decoding trajectory. The same pattern holds in the multi-topic setting containing 60 questions from 57 MMLU subjects. The overlap remains approximately $0.75$ under the pair-norm score and approximately $0.71$ under the absolute pair dot product across all four decoding ranges. The overlap therefore remains stable not only under long generation but also when the prompt spans many different topics. Across both stress-test settings, the $\tilde{Q}\tilde{K}$ pair-norm criterion also maintains the highest overlap among the compared selection criteria. The small differences between the maximum and minimum overlap further confirm limited temporal variation. A similar pattern at $r=10\%$ shows that this behavior is not specific to the $r=50\%$ selection ratio.

Overall, these results indicate that the overlap between the channel pairs selected during prefill and decoding-time salient channel pairs remains relatively stable throughout decoding in the tested long-generation and multi-topic settings.

\section{Use of AI Assistants}
AI assistants were used only for language editing, including grammar and spelling checks and synonym suggestions, and for assistance with Triton code implementation.
All technical content and implementation details were verified and finalized by the authors.

\begin{table*}[t]
  \centering
  \small
  \begin{tabular}{ll|>{\columncolor{gray!20}}c|cccc}
    \toprule
    Model & Task & K16V16 & K2.25V16 & K2.25*V16 & K16V2.25 & K16V2.25* \\
    \midrule
    \multirow{2}{*}{R1-Llama-8B}
      & AIME24 & $48.33_{\pm 6.19}$ & $28.67_{\pm 7.02}$ & $44.00_{\pm 7.27}$ & $40.67_{\pm 5.33}$ & $\mathbf{51.00_{\pm 6.51}}$ \\
      & AIME25 & $31.33_{\pm 8.19}$ & $26.67_{\pm 5.16}$ & $28.67_{\pm 3.71}$ & $26.33_{\pm 4.58}$ & $\mathbf{32.00_{\pm 5.81}}$ \\
    \midrule
    \multirow{2}{*}{Qwen3-8B}
      & AIME24 & $76.33_{\pm 4.82}$ & $64.00_{\pm 5.54}$ & $70.00_{\pm 2.98}$ & $68.33_{\pm 4.53}$ & $\mathbf{73.33_{\pm 2.98}}$ \\
      & AIME25 & $67.00_{\pm 4.58}$ & $55.33_{\pm 4.99}$ & $59.67_{\pm 4.07}$ & $60.00_{\pm 2.11}$ & $\mathbf{68.00_{\pm 5.42}}$ \\
    \bottomrule
  \end{tabular}
  \caption{\textbf{Sensitivity of key/value caches to quantization.}
  Each cell reports mean $\pm$ std over 10 responses per question.
  All quantized methods use KIVI~\citep{kivi} ($G=128$, $N=128$); method names ending with (*) preserve the first 32 tokens in full precision.
  The best quantized method per row within each model block is in \textbf{bold}.}
  \label{appendix:kv-sink-compare}
\end{table*}

\begin{table*}[t]
\centering
\small
\setlength{\tabcolsep}{5pt}
\begin{tabular}{lcccc}
\toprule
\multirow{2}{*}{Method} & \multicolumn{2}{c}{R1-Qwen-7B (AIME25)} & \multicolumn{2}{c}{R1-Llama-8B (LiveCodeBench~v5)} \\
\cmidrule(lr){2-3} \cmidrule(lr){4-5}
 & $\Delta_{\tilde{K}}$ & $D_{\mathrm{KL}}^{A}$ & $\Delta_{\tilde{K}}$ & $D_{\mathrm{KL}}^{A}$ \\
\midrule
\multicolumn{5}{l}{\textit{Only 3-bit (Uniform precision)}} \\
\quad Per-Channel                                & 0.042 & 0.043 & 0.058 & 0.020 \\
\quad Per-Channel after $(r,\theta)$             & 0.098 & 0.101 & 0.142 & 0.043 \\
\midrule
\multicolumn{5}{l}{\textit{Only 2-bit (Uniform precision)}} \\
\quad Per-Channel                                & 0.230 & 0.232 & 0.324 & 0.120 \\
\quad Per-Channel after $(r,\theta)$             & 0.499 & 0.518 & 0.722 & 0.213 \\
\midrule
\multicolumn{5}{l}{\textit{H3(50\%)L1 (Full token based)}} \\
\quad $\tilde{K}$ range                  & $0.184_{(0.913)}$ & $0.125_{(0.378)}$ & $0.286_{(1.453)}$ & $0.094_{(0.330)}$ \\
\quad $\tilde{K}$ magnitude              & $0.194_{(0.975)}$ & $0.087_{(0.294)}$ & $0.307_{(1.588)}$ & $0.065_{(0.294)}$ \\
\quad $\tilde{K}$ pair norm              & $0.194_{(0.976)}$ & $0.083_{(0.291)}$ & $0.313_{(1.629)}$ & $0.060_{(0.293)}$ \\
\quad $\tilde{Q}\tilde{K}$ pair norm     & $0.197_{(0.998)}$ & $0.080_{(0.273)}$ & $0.320_{(1.668)}$ & $0.055_{(0.279)}$ \\
\midrule
\multicolumn{5}{l}{\textit{H3(50\%)L1 (Prefill token based)}} \\
\quad $\tilde{K}$ range                  & $0.193_{(0.966)}$ & $0.128_{(0.473)}$ & $0.281_{(1.430)}$ & $0.078_{(0.307)}$ \\
\quad $\tilde{K}$ magnitude              & $0.198_{(0.998)}$ & $0.096_{(0.362)}$ & $0.303_{(1.561)}$ & $0.062_{(0.299)}$ \\
\quad $\tilde{K}$ pair norm              & $0.196_{(0.988)}$ & $0.090_{(0.334)}$ & $0.309_{(1.602)}$ & $0.059_{(0.293)}$ \\
\quad $\tilde{Q}\tilde{K}$ pair norm     & $0.201_{(1.022)}$ & $0.082_{(0.283)}$ & $0.317_{(1.649)}$ & $0.055_{(0.278)}$ \\
\bottomrule
\end{tabular}
\caption{\textbf{Key-cache quantization simulation.} Table shows $\Delta_{\tilde{K}}$ (key MSE) and $D_{\mathrm{KL}}^{A}$ (attention KL divergence) for uniform-precision and H3(50\%)L1 mixed-precision methods on R1-Qwen-7B (AIME25) and R1-Llama-8B (LiveCodeBench~v5). All methods use per-channel asymmetric quantization ($G=128$); 1-bit groups use the correction of Eq.~\ref{eq:1bit_correction}. Each H3(50\%)L1 entry reports the value \emph{with} 1-bit correction, with the value \emph{without} correction shown as a parenthesized subscript. Mixed-precision rows differ only in the channel(-pair) importance score used to allocate the high/low bits, with importance computed over either the full sequence or prefill tokens only.}
\label{appendix:k-quant-simul-table}
\end{table*}

\begin{table*}[t]
\centering
\small
\setlength{\tabcolsep}{4pt}
\renewcommand{\arraystretch}{1.25}

\begin{tabular}{lcccccccc}
\toprule
\multicolumn{9}{l}{\textit{(a) Method configurations}} \\
\midrule

\textbf{Method}
& $b$
& $b_H$
& $b_L$
& $r$ / $r_h$
& $r_{\mathrm{eff}}$
& $G$
& $N$
& $N_s$ \\
\midrule

KIVI
& 2
& --
& --
& --
& --
& 128
& 128
& 0 \\

PolarQuant
& 2
& --
& --
& --
& --
& 128
& 128
& 0 \\

ChanMix
& --
& 4
& 3, 2, 1
& 0.1
& --
& 128
& 128
& 0 \\

Kitty
& --
& 4
& 2
& 0.03125
& $\tfrac{\lfloor d r \rfloor}{d}$
& 128
& 128
& 32 \\

PN-Mix (1)
& --
& 3
& 1
& 0.5
& $\tfrac{2\lfloor r d/2 \rfloor}{d}$
& 128
& 128
& 32 \\

PN-Mix (2)
& --
& 3
& 1
& 0.4
& $\tfrac{2\lfloor r d/2 \rfloor}{d}$
& 64
& 128
& 32 \\

\bottomrule
\end{tabular}

\vspace{6pt}

\begin{tabular}{llcccc}
\toprule
\multicolumn{6}{l}{\textit{(b) Average key-cache bit-width accounting}} \\
\midrule

\textbf{Method}
& \makecell{\textbf{Effective K-storage}\\\textbf{formula}}
& \makecell{\textbf{Effective K}\\\textbf{storage bits}}
& \makecell{\textbf{Saliency metadata}\\\textbf{formula}}
& \makecell{\textbf{Avg. metadata}\\\textbf{overhead}}
& \makecell{\textbf{Total K}\\\textbf{avg. bit}} \\
\midrule

KIVI
& $b+\tfrac{32}{G}$
& 2.2500
& --
& 0
& 2.2500 \\

PolarQuant
& $b+\tfrac{32}{G}$
& 2.2500
& --
& 0
& 2.2500 \\

ChanMix
& $
2\left(
1+
\tfrac{\lfloor L H^{kv} r_h\rfloor}{L H^{kv}}
\right)
+
\tfrac{16}{G}
$
& 2.3203
& $\tfrac{64}{N_s+T_q}$
& $\approx 0$
& 2.3203 \\

Kitty
& $
\tfrac{
16N_s+
T_q\left[
r_{\mathrm{eff}}b_H+
(1-r_{\mathrm{eff}})b_L+
\tfrac{32}{G}
\right]
}{
N_s+T_q
}
$
& 2.3259
& $\tfrac{8T_q}{G(N_s+T_q)}$
& 0.0624
& 2.3883 \\

PN-Mix (1)
& $
\tfrac{
16N_s+
T_q\left[
r_{\mathrm{eff}}b_H+
(1-r_{\mathrm{eff}})b_L+
\tfrac{32}{G}
\right]
}{
N_s+T_q
}
$
& 2.2634
& $\tfrac{1}{2(N_s+T_q)}$
& $\approx 0$
& 2.2634 \\

PN-Mix (2)
& $
\tfrac{
16N_s+
T_q\left[
r_{\mathrm{eff}}b_H+
(1-r_{\mathrm{eff}})b_L+
\tfrac{32}{G}
\right]
}{
N_s+T_q
}
$
& 2.2946
& $\tfrac{1}{2(N_s+T_q)}$
& $\approx 0$
& 2.2946 \\

\bottomrule
\end{tabular}
\caption{\textbf{Method configurations and average key-cache bit-widths.}
Average bit-widths are computed with $T_q=32{,}768$, where $T_q$ denotes the number of quantized tokens; the FP16 residual buffer is excluded from the calculation.
For Kitty and PN-Mix, $r_{\mathrm{eff}}$ denotes the actual fraction of channels assigned to high precision after converting the nominal ratio $r$ to an integer number of selected channels or RoPE channel pairs, respectively.
Effective K-storage bits exclude saliency metadata, which is reported separately and added to obtain the total average K bit-width.}
\label{tab:bitwidth_config_summary}
\end{table*}

\begin{table*}[t]
\centering
\small
\setlength{\tabcolsep}{3.5pt}

\begin{tabular}{llccccc}
\toprule
\multicolumn{7}{l}{
\textit{(a) Long generation: R1-Qwen-7B / AIME25,
96 prefill tokens + 32,672 decoding steps}
} \\
\midrule
\textbf{Reference score}
& \textbf{Selection criterion}
& \textbf{0--8K}
& \textbf{8--16K}
& \textbf{16--24K}
& \textbf{24K--32.7K}
& \textbf{Max--Min} \\
\midrule

\multirow{4}{*}{$\|\mathbf{q}_i^t\|_2\|\mathbf{k}_i^s\|_2$}
& $\tilde{K}$ range
& $0.61/0.35$
& $0.62/0.35$
& $0.61/0.35$
& $0.61/0.35$
& $0.009/0.029$ \\

& $\tilde{K}$ magnitude
& $0.68/0.49$
& $0.69/0.50$
& $0.69/0.50$
& $0.69/0.51$
& $0.008/0.034$ \\

& $\tilde{K}$ pair norm
& $0.70/0.54$
& $0.70/0.56$
& $0.70/0.56$
& $0.70/0.56$
& $0.010/0.040$ \\

& \textbf{$\tilde{Q}\tilde{K}$ pair norm}
& $\mathbf{0.72/0.62}$
& $\mathbf{0.72/0.61}$
& $\mathbf{0.72/0.61}$
& $\mathbf{0.72/0.61}$
& $\mathbf{0.007/0.027}$ \\

\midrule

\multirow{4}{*}{$|\langle \mathbf{q}_i^t,\mathbf{k}_i^s\rangle|$}
& $\tilde{K}$ range
& $0.59/0.33$
& $0.60/0.33$
& $0.60/0.33$
& $0.60/0.33$
& $0.008/0.024$ \\

& $\tilde{K}$ magnitude
& $0.65/0.45$
& $0.65/0.46$
& $0.65/0.46$
& $0.65/0.46$
& $0.007/0.023$ \\

& $\tilde{K}$ pair norm
& $0.66/0.50$
& $0.67/0.51$
& $0.67/0.50$
& $0.67/0.51$
& $0.008/0.026$ \\

& \textbf{$\tilde{Q}\tilde{K}$ pair norm}
& $\mathbf{0.68/0.56}$
& $\mathbf{0.68/0.55}$
& $\mathbf{0.68/0.54}$
& $\mathbf{0.68/0.54}$
& $\mathbf{0.006/0.029}$ \\

\bottomrule
\end{tabular}

\vspace{6pt}

\begin{tabular}{llccccc}
\toprule
\multicolumn{7}{l}{
\textit{(b) Multi-topic generation: Llama-3.1-8B-Instruct /
LongGenBench-MMLU, 10,366 prefill tokens + 3,970 decoding steps}
} \\
\midrule
\textbf{Reference score}
& \textbf{Selection criterion}
& \textbf{0--992}
& \textbf{992--1984}
& \textbf{1984--2976}
& \textbf{2976--3970}
& \textbf{Max--Min} \\
\midrule

\multirow{4}{*}{$\|\mathbf{q}_i^t\|_2\|\mathbf{k}_i^s\|_2$}
& $\tilde{K}$ range
& $0.65/0.30$
& $0.66/0.30$
& $0.65/0.29$
& $0.65/0.29$
& $0.001/0.007$ \\

& $\tilde{K}$ magnitude
& $0.71/0.55$
& $0.71/0.55$
& $0.71/0.56$
& $0.71/0.56$
& $0.002/0.007$ \\

& $\tilde{K}$ pair norm
& $0.73/0.68$
& $0.73/0.68$
& $0.73/0.68$
& $0.73/0.68$
& $0.002/0.008$ \\

& \textbf{$\tilde{Q}\tilde{K}$ pair norm}
& $\mathbf{0.75/0.72}$
& $\mathbf{0.75/0.72}$
& $\mathbf{0.75/0.72}$
& $\mathbf{0.75/0.73}$
& $\mathbf{0.002/0.009}$ \\

\midrule

\multirow{4}{*}{$|\langle \mathbf{q}_i^t,\mathbf{k}_i^s\rangle|$}
& $\tilde{K}$ range
& $0.63/0.27$
& $0.63/0.27$
& $0.63/0.28$
& $0.63/0.28$
& $0.002/0.004$ \\

& $\tilde{K}$ magnitude
& $0.67/0.46$
& $0.68/0.47$
& $0.68/0.47$
& $0.68/0.48$
& $0.003/0.020$ \\

& $\tilde{K}$ pair norm
& $0.69/0.56$
& $0.69/0.56$
& $0.69/0.57$
& $0.69/0.58$
& $0.003/0.023$ \\

& \textbf{$\tilde{Q}\tilde{K}$ pair norm}
& $\mathbf{0.71/0.58}$
& $\mathbf{0.71/0.59}$
& $\mathbf{0.71/0.60}$
& $\mathbf{0.71/0.61}$
& $\mathbf{0.003/0.024}$ \\

\bottomrule
\end{tabular}
\caption{\textbf{Stability of prefill-selected channel pairs during decoding.}
We measure the overlap between the fixed top-$r\%$ channel pairs selected from prefill and the top-$r\%$ channel pairs identified by a step-wise reference during decoding.
The reference scores are $\|\mathbf{q}_i^t\|_2\|\mathbf{k}_i^s\|_2$ and $|\langle \mathbf{q}_i^t,\mathbf{k}_i^s\rangle|$.
For each decode query at step $t$, the reference considers only decode-generated causal keys ($P \le s \le t$), where $P$ is the prefill length.
Each cell reports the overlap ratio aggregated over all eligible query--key interactions, heads, and layers within the bin, in the order $r=50\% / r=10\%$.
Overlap ranges from 0 to 1, where 1 indicates identical selections.
Max--Min is computed over 1K-step bins for the long-generation setting and 992-step bins for the multi-topic setting.}
\label{tab:prefill_drift_decode}
\end{table*}

\begin{table*}[t]
\centering
\small
\setlength{\tabcolsep}{4.5pt}
\renewcommand{\arraystretch}{0.95}
\begin{tabular}{lccccccc}
\toprule
Method & K bits & KV-Retr. & LongDial.-QA & LongBook-Choice & LongBook-QA & LongBook-Sum & Avg. \\
 & & (acc) & (acc) & (acc) & (F1) & (ROUGE-L-Sum) & \\
\midrule
\rowcolor{gray!15}
FP16 & 16 & $\mathrm{59.00}$ & $\mathrm{11.00}$ & $\mathrm{62.00}$ & $\mathrm{9.66}$ & $\mathrm{29.07}$ & $\mathrm{34.15}$ \\
KIVI        & 2.25 & $\mathrm{32.00}$ & $\mathrm{9.00}$ & $\mathrm{58.00}$ & $\mathbf{10.03}$ & $\mathrm{25.55}$ & $\mathrm{26.92}$ \\
ChanMix     & 2.32 & $\mathrm{53.00}$ & $\mathrm{9.00}$ & $\mathrm{58.00}$ & $\mathrm{10.01}$ & $\mathrm{26.74}$ & $\mathrm{31.35}$ \\
Kitty       & 2.39 & $\mathrm{49.00}$ & $\mathbf{10.00}$ & $\mathrm{61.00}$ & $\mathrm{8.82}$ & $\mathbf{29.09}$ & $\mathrm{31.58}$ \\
PN-Mix (1)  & 2.26 & $\mathbf{59.00}$ & $\mathrm{7.00}$ & $\mathrm{62.00}$ & $\mathrm{8.67}$ & $\mathrm{28.76}$ & $\mathrm{33.09}$ \\
PN-Mix (2)  & 2.29 & $\mathbf{59.00}$ & $\mathbf{10.00}$ & $\mathbf{63.00}$ & $\mathrm{9.14}$ & $\mathrm{28.77}$ & $\mathbf{33.98}$ \\
\bottomrule
\end{tabular}
\caption{\textbf{Accuracy comparison on InfiniteBench with Llama-3.1-8B-Instruct.}
We evaluate the first 100 examples of each of five English tasks.
Prompts are built with the official task templates and middle-truncated to
32{,}000 tokens using the official truncation routine; no chat template is
applied, and decoding is greedy with the official per-task generation budgets
(6--1{,}200 tokens).
All compressed configurations (i.e., every method except FP16) are matched to
an average key-cache precision of 2.25--2.39 bits, while the value cache is kept in full precision.
Bold indicates the best non-FP16 result in each column.}
\label{appendix:infinitebench}
\end{table*}

\begin{table*}[t]
  \centering
  \small
  \setlength{\tabcolsep}{3pt}
  \renewcommand{\arraystretch}{0.95}
  \begin{tabular}{ll|ccccccccccccc|c}
    \toprule
    Len. & Method & S1 & S2 & S3 & MK1 & MK2 & MK3 & MV & MQ & FWE & CWE & QA1 & QA2 & VT & Avg. \\
    \midrule
    \multirow{6}{*}{16K}
      & \cellcolor{gray!15}FP16       & \cellcolor{gray!15}100.0 & \cellcolor{gray!15}100.0 & \cellcolor{gray!15}100.0 & \cellcolor{gray!15}99.5 & \cellcolor{gray!15}100.0 & \cellcolor{gray!15}100.0 & \cellcolor{gray!15}100.0 & \cellcolor{gray!15}99.4 & \cellcolor{gray!15}95.2 & \cellcolor{gray!15}86.6 & \cellcolor{gray!15}70.8 & \cellcolor{gray!15}57.5 & \cellcolor{gray!15}99.8 & \cellcolor{gray!15}92.98 \\
      & KIVI       & 100.0 & 99.5 & 93.5 & 99.5 & 96.5 & 69.0 & 95.2 & 95.4 & 93.0 & 84.7 & 71.5 & 54.5 & 95.2 & 88.26 \\
      & ChanMix    & 100.0 & 99.5 & 99.0 & 99.0 & 99.0 & 97.5 & 97.2 & 97.5 & 92.7 & 87.7 & 70.8 & 56.5 & 98.7 & 91.93 \\
      & Kitty      & 100.0 & 100.0 & 99.5 & 99.5 & 98.5 & 95.0 & 99.4 & 98.5 & 94.3 & 87.1 & 72.8 & 54.5 & 98.1 & 92.09 \\
      & PN-Mix (1) & 100.0 & 100.0 & 100.0 & 100.0 & 100.0 & 99.0 & 100.0 & 99.4 & 95.7 & 85.0 & 72.6 & 56.5 & 99.4 & \textbf{92.89} \\
      & PN-Mix (2) & 100.0 & 100.0 & 100.0 & 99.5 & 99.5 & 100.0 & 99.9 & 99.1 & 95.0 & 82.2 & 72.1 & 56.5 & 99.4 & 92.55 \\
    \midrule
    \multirow{6}{*}{32K}
      & \cellcolor{gray!15}FP16       & \cellcolor{gray!15}100.0 & \cellcolor{gray!15}100.0 & \cellcolor{gray!15}100.0 & \cellcolor{gray!15}100.0 & \cellcolor{gray!15}99.5 & \cellcolor{gray!15}99.0 & \cellcolor{gray!15}99.4 & \cellcolor{gray!15}98.8 & \cellcolor{gray!15}93.2 & \cellcolor{gray!15}22.8 & \cellcolor{gray!15}74.6 & \cellcolor{gray!15}53.0 & \cellcolor{gray!15}98.9 & \cellcolor{gray!15}87.62 \\
      & KIVI       & 100.0 & 100.0 & 95.0 & 99.5 & 95.0 & 47.5 & 94.9 & 92.4 & 91.3 & 28.6 & 69.4 & 49.5 & 94.2 & 81.32 \\
      & ChanMix    & 99.5 & 100.0 & 99.0 & 100.0 & 98.5 & 91.5 & 96.8 & 98.5 & 92.7 & 21.7 & 71.9 & 52.0 & 96.7 & 86.06 \\
      & Kitty      & 100.0 & 100.0 & 98.5 & 100.0 & 97.0 & 82.0 & 96.6 & 98.8 & 92.5 & 34.5 & 71.8 & 51.0 & 97.6 & 86.17 \\
      & PN-Mix (1) & 100.0 & 100.0 & 100.0 & 100.0 & 99.0 & 98.5 & 97.8 & 98.9 & 93.5 & 21.6 & 74.4 & 51.5 & 99.0 & \textbf{87.24} \\
      & PN-Mix (2) & 100.0 & 99.0 & 100.0 & 100.0 & 99.0 & 98.5 & 98.2 & 98.2 & 93.8 & 16.7 & 73.9 & 53.0 & 97.8 & 86.79 \\
    \midrule
    \multirow{6}{*}{64K}
      & \cellcolor{gray!15}FP16       & \cellcolor{gray!15}100.0 & \cellcolor{gray!15}100.0 & \cellcolor{gray!15}100.0 & \cellcolor{gray!15}99.5 & \cellcolor{gray!15}98.0 & \cellcolor{gray!15}98.5 & \cellcolor{gray!15}98.4 & \cellcolor{gray!15}99.8 & \cellcolor{gray!15}87.5 & \cellcolor{gray!15}0.6 & \cellcolor{gray!15}68.8 & \cellcolor{gray!15}50.0 & \cellcolor{gray!15}96.8 & \cellcolor{gray!15}84.44 \\
      & KIVI       & 99.0 & 99.5 & 94.5 & 98.0 & 91.0 & 22.5 & 91.9 & 96.4 & 86.8 & 1.8 & 65.6 & 48.5 & 90.6 & 75.85 \\
      & ChanMix    & 100.0 & 99.5 & 99.5 & 99.5 & 98.0 & 76.0 & 93.1 & 98.6 & 87.3 & 1.7 & 66.4 & 48.0 & 92.3 & 81.54 \\
      & Kitty      & 100.0 & 99.5 & 97.5 & 99.5 & 94.5 & 58.5 & 97.4 & 99.4 & 87.8 & 2.2 & 67.6 & 49.0 & 91.2 & 80.32 \\
      & PN-Mix (1) & 100.0 & 100.0 & 100.0 & 99.5 & 97.0 & 98.0 & 97.4 & 99.5 & 88.2 & 0.9 & 68.9 & 50.5 & 94.1 & \textbf{84.15} \\
      & PN-Mix (2) & 100.0 & 100.0 & 100.0 & 99.5 & 97.5 & 98.5 & 97.0 & 99.9 & 88.2 & 0.8 & 67.8 & 50.5 & 93.4 & 84.08 \\
    \bottomrule
  \end{tabular}
  \caption{\textbf{Accuracy on RULER across context lengths from 16K to 64K with Llama-3.1-8B-Instruct.}
  Tasks are the 13 official RULER categories, where NIAH denotes needle-in-a-haystack:
  single NIAH (S1--S3), multi-key NIAH (MK1--MK3), multi-value NIAH (MV), multi-query NIAH (MQ),
  frequent-word extraction (FWE), common-word extraction (CWE),
  question answering (QA1--QA2), and variable tracking (VT).
  All compressed configurations (i.e., every method except FP16) are matched to
  an average key-cache precision of 2.25--2.39 bits, while the value cache is kept in full precision.
  Bold indicates the best non-FP16 result in each column.}
  \label{appendix:ruler}
\end{table*}

\begin{table*}[t]
\centering
\small
\setlength{\tabcolsep}{5pt}
\begin{tabular}{lcccc}
\toprule
\multirow{2}{*}{Configuration} & \multirow{2}{*}{Avg key bit} & \multicolumn{2}{c}{AIME24} \\
\cmidrule(lr){3-4}
 & & Qwen3-8B & R1-Llama-8B \\
\midrule
\textit{Full} (FP16)                                                            & 16.00 & $\mathrm{76.33}_{\pm 4.82}$ & $\mathrm{48.33}_{\pm 6.19}$ \\
\midrule
\multicolumn{4}{l}{\textit{KIVI}} \\
\quad $b{=}2,\,G{=}32$                                                          & 3.00 & $\mathrm{72.33}_{\pm 3.96}$ & $\mathrm{42.67}_{\pm 8.14}$ \\
\quad $b{=}2,\,G{=}64$                                                          & 2.50 & $\mathrm{68.00}_{\pm 4.99}$ & $\mathrm{35.00}_{\pm 9.80}$ \\
\quad $b{=}2,\,G{=}128$                                                         & 2.25 & $\mathrm{64.00}_{\pm 5.54}$ & $\mathrm{28.67}_{\pm 7.02}$ \\
\quad $b{=}1,\,G{=}32$                                                          & 2.00 & $\mathrm{0.00}_{\pm 0.00}$  & $\mathrm{0.00}_{\pm 0.00}$  \\
\quad $b{=}1,\,G{=}64$                                                          & 1.50 & $\mathrm{0.00}_{\pm 0.00}$  & $\mathrm{0.00}_{\pm 0.00}$  \\
\midrule
\multicolumn{4}{l}{\textit{Kitty}} \\
\quad $b_H{=}4,\,b_L{=}2,\,r{=}30\%,\,G{=}128$                                  & 2.92 & $\mathrm{77.67}_{\pm 4.96}$ & $\mathrm{48.33}_{\pm 7.19}$ \\
\quad $b_H{=}4,\,b_L{=}2,\,r{=}10\%,\,G{=}128$                                  & 2.51 & $\mathrm{73.33}_{\pm 4.71}$ & $\mathrm{46.67}_{\pm 3.94}$ \\
\quad $b_H{=}4,\,b_L{=}2,\,r{=}3.125\%,\,G{=}128$                               & 2.39 & $\mathrm{70.67}_{\pm 5.93}$ & $\mathrm{42.00}_{\pm 5.62}$ \\
\quad $b_H{=}3,\,b_L{=}1,\,r{=}30\%,\,G{=}64$                                   & 2.23 & $\mathrm{40.00}_{\pm 6.50}$ & $\mathrm{1.33}_{\pm 1.63}$  \\
\quad $b_H{=}3,\,b_L{=}1,\,r{=}20\%,\,G{=}64$                                   & 2.03 & $\mathrm{9.33}_{\pm 2.49}$  & $\mathrm{0.00}_{\pm 0.00}$  \\
\quad $b_H{=}3,\,b_L{=}1,\,r{=}10\%,\,G{=}128$                                  & 1.51 & $\mathrm{0.00}_{\pm 0.00}$  & $\mathrm{0.00}_{\pm 0.00}$  \\
\midrule
\multicolumn{4}{l}{\textit{PN-Mix (Ours)}} \\
\quad $b_H{=}3,\,b_L{=}2,\,r{=}70\%,\,G{=}128$                                  & 2.95 & $\mathrm{76.67}_{\pm 3.33}$ & $\mathrm{48.33}_{\pm 7.19}$ \\
\quad $b_H{=}3,\,b_L{=}2,\,r{=}40\%,\,G{=}64$                                   & 2.90 & $\mathrm{75.33}_{\pm 4.52}$ & $\mathrm{48.67}_{\pm 5.62}$ \\
\quad $b_H{=}4,\,b_L{=}1,\,r{=}30\%,\,G{=}64$                                   & 2.40 & $\mathrm{75.00}_{\pm 5.00}$ & $\mathrm{48.33}_{\pm 6.19}$ \\
\quad $b_H{=}3,\,b_L{=}1,\,r{=}40\%,\,G{=}64$                                   & 2.29 & $\mathrm{76.00}_{\pm 4.67}$ & $\mathrm{50.67}_{\pm 5.54}$ \\
\quad $b_H{=}3,\,b_L{=}1,\,r{=}40\%,\,G{=}128$                                  & 2.04 & $\mathrm{74.67}_{\pm 3.40}$ & $\mathrm{47.33}_{\pm 6.11}$ \\
\quad $b_H{=}3,\,b_L{=}1,\,r{=}20\%,\,G{=}64$                                   & 1.89 & $\mathrm{75.00}_{\pm 3.07}$ & $\mathrm{46.67}_{\pm 3.65}$ \\
\quad $b_H{=}3,\,b_L{=}1,\,r{=}10\%,\,G{=}128$                                  & 1.45 & $\mathrm{67.66}_{\pm 3.95}$  & $\mathrm{40.33}_{\pm 4.58}$ \\
\bottomrule
\end{tabular}
\caption{\textbf{Configurations for the bit-rate Pareto frontier in Figure~\ref{fig:pareto}.}
For each operating point, we report the selected hyperparameters and the corresponding AIME24 accuracy (\%) with $b_V=16$ fixed.
Kitty and PN-Mix both preserve $N_s=32$ sink tokens in FP16.
\textit{Full} denotes the FP16 KV-cache baseline.}
\label{appendix:pareto-config-table}
\end{table*}

\end{document}